\documentclass[preprint,12pt]{elsarticle}
\usepackage{amssymb}
\usepackage{amsmath}
\usepackage{dsfont}
\usepackage[linesnumbered,ruled,vlined]{algorithm2e}
\usepackage{amsmath}
\usepackage{graphicx}
\usepackage{subcaption}
\usepackage{lineno}
\usepackage{amssymb}
\usepackage{amsthm}
\theoremstyle{definition}
\usepackage{graphics} % for pdf, bitmapped graphics files
\usepackage{epsfig} % for postscript graphics files
\usepackage{geometry}
\usepackage{hyperref}
\usepackage{algpseudocode}
\usepackage{xcolor}
\usepackage{booktabs}
\RestyleAlgo{ruled}
\DeclareMathOperator*{\argmax}{arg\,max}

\DeclareMathAlphabet{\mathcal}{OMS}{cmsy}{m}{n} % from https://tex.stackexchange.com/a/109971/201916

\journal{Artificial Intelligence}

\begin{document}
\newtheorem*{definition}{Definition}

\begin{frontmatter}



\title{Partner Capability Estimation for Task-Agnostic Adaptation in Ad-Hoc Teamwork}


\author[kcl,ox]{Peter Tisnikar\fnref{note}\corref{cor}} %% Author name
\author[kcl]{Maja Swieczkowska}
\author[kcl]{Benteng Ma}
\author[kcl]{Gerard Canal}
\author[kcl]{Matteo Leonetti\corref{cor}}

\affiliation[kcl]{organization={Department of Informatics, King's College London},%Department and Organization
            addressline={30 Aldwych}, 
            city={London},
            postcode={WC2B 4BG}, 
            state={England},
            country={United Kingdom}}
            
\affiliation[ox]{organization={Department of Engineering Science, University of Oxford},
            addressline={Parks Road},
            city={Oxford},
            postcode={OX1 3PJ},
            state={England},
            country={United Kingdom}
            }
\fntext[note]{Work done whilst at King's College London. This work was supported by UK Research and Innovation (grant number EP/S023356/1), in the UKRI Centre for Doctoral Training in Safe and Trusted Artificial Intelligence (\texttt{www.safeandtrustedai.org}).}
\cortext[cor]{Corresponding author.}

%% Abstract
\begin{abstract}
Effective collaboration with novel and diverse partners is a crucial skill for autonomous agents. Most current ad-hoc teamwork (AHT) approaches assume that agents will collaborate on a single, fixed task and that the partner's capabilities, their ability to successfully execute the desired action, are already known. In reality, a partner's true capabilities are often hidden, and human collaborators may act sub-optimally on tasks with multiple valid strategies. To address these limitations, we extend ad-hoc teamwork into a multi-task setting by re-framing it as a problem of joint planning with decentralised execution under hidden partner capabilities. We introduce \textbf{CE-CM} (\textbf{C}apability \textbf{E}stimation via \textbf{C}ontextual \textbf{M}odels), an approximate Bayesian method that infers task-invariant capability vectors. By using simulation-based sampling, the agent estimates capabilities and induces a contextual Multi-agent Markov Decision Processes for planning. This approach requires no population pre-training and refines its beliefs online from just a few tasks. To account for human unpredictability, we propose \textbf{CE-CM-Div}, an extension that evaluates capability hypotheses against diverse planner rollouts rather than a single optimal trajectory. Simulated experiments demonstrate that CE-CM rapidly recovers hidden capabilities, reduces infeasible action assignments, and adapts to changes over time. Furthermore, in an offline human study of 225 trajectories from 15 participants, CE-CM-Div substantially improved capability estimates over the baseline CE-CM method. Our results suggest capability-based modelling is a promising interpretable, task-agnostic representation in the studied settings, demonstrating that accounting for behavioural diversity is essential for robust human-AI teaming.
\end{abstract}



%% Keywords

\end{frontmatter}


\section{Introduction}

Autonomous agents will increasingly be required to collaborate with diverse partners. In open-world settings, an agent cannot assume that its partners will be homogeneous; they may differ in their preferences, skill level, physical or cognitive capabilities, and other traits. Furthermore, these traits may only be revealed to the agent through repeated interaction. Adapting to such collaborators, especially human ones, has therefore been recognised as a crucial requirement for intelligent embodied agents \cite{Zhao2022the, natarajan2023HumanRobot}. Consider, for example, a household robot collaborating with a new human partner on a series of chores. If the human leaves the robot to carry a bulky object or repeatedly avoids part of the environment, such behaviour is difficult to interpret: it may reflect a deliberate strategy, but it may also indicate that the partner is simply unable to perform that part of the task. This distinction is important for coordination, because it changes which actions should be assigned to which agent and which joint plans are likely to succeed. We argue that such adaptation requires learning an explicit, reusable model of the current partner, rather than only a policy that is robust to partner variation within a single task.

Ad-hoc teamwork (AHT)~\cite{mirsky2022Survey, stone2010Ad} provides a natural framework for this problem, as it studies agents that must collaborate with previously unknown partners without pre-coordination. However, much recent AHT work learns robust ego policies against populations of partners in a fixed task, often using model-free Multi-agent reinforcement learning (see Section \ref{sec:rw} for a more detailed discussion). Such policies can generalise within the training task, but they usually do not provide an explicit model of the current partner, nor do they transfer directly across tasks. We therefore focus on repeated collaboration, where the agent must learn something about the particular partner and reuse that knowledge in future tasks.

This leads us to the central question of the paper: what kind of partner representation can be learned from interaction, reused across tasks, and used to adapt to the current individual? We argue that this requires shifting the focus of AHT toward transferable partner models. In particular, we propose to represent partners through their \emph{capabilities}: latent, task-invariant constraints on their ability to realise transitions in a joint task. Whereas preferences and goals describe what a partner may want to do, capabilities describe what they are fundamentally able to do. Policies are tied to particular tasks and observed actions are local to particular states, while capabilities can explain a broader family of feasible and infeasible interactions across different tasks. By modelling what the partner can and cannot do, rather than only what they happened to do before, the agent can accumulate knowledge about an individual partner and use it to adapt its own behaviour in future tasks.

If the partner is modelled in terms of hidden capabilities, adaptation becomes a problem of reasoning about how those capabilities constrain feasible joint behaviour. In cooperative settings with shared goals, the agent must infer which joint plans are possible under different assumptions about what the partner can and cannot do, and allocate work accordingly. This leads naturally to viewing the interaction as joint planning with decentralised execution under hidden partner capabilities (Section~\ref{sec:problem}): the team’s behaviour is reasoned about jointly, but each agent ultimately executes only its own actions. Under this view, the same capability-conditioned model supports both explanation and adaptation: observed trajectories reveal which capability assignments best explain prior interactions, and the resulting estimate is used to plan the next task with the same partner.

\begin{figure}[t!]
    \centering
    \includegraphics[width=0.99\linewidth, trim={0.8cm 6cm 1.2cm 7cm},clip]{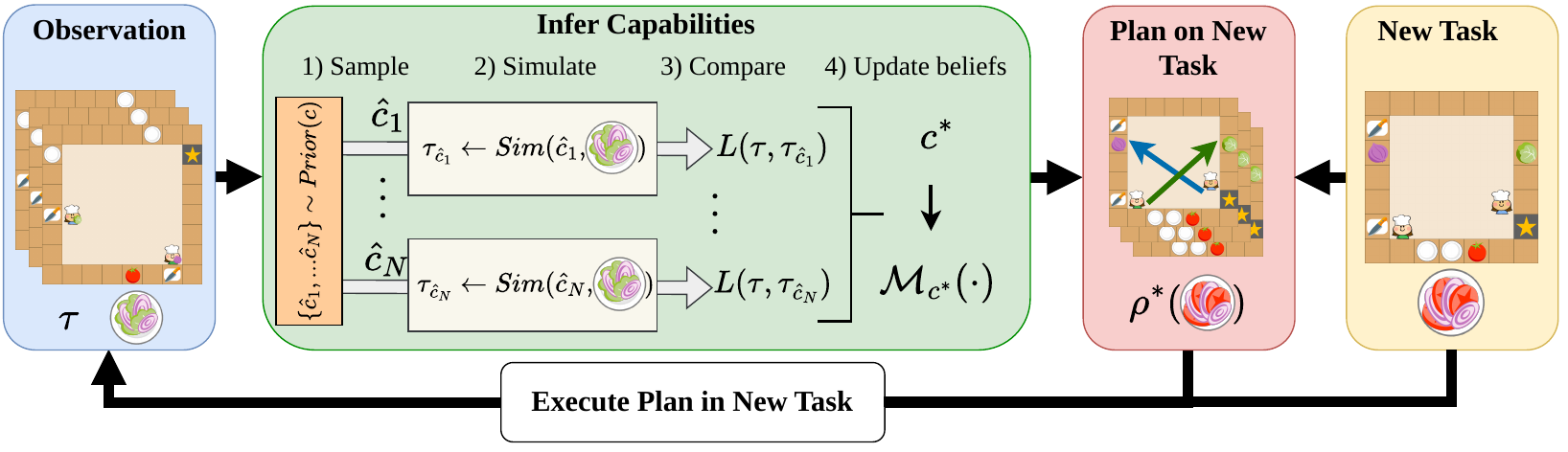}
    \caption{Overview of CE-CM. After completing a task with a partner, the agent uses the observed joint trajectory and the task's known goal to infer what the partner is capable of doing. It samples candidate capability vectors $\hat{c}_1, \dots, \hat{c}_N$, simulates the joint trajectory that each would induce in the observed task, and compares the simulated trajectories with the observation. Capability hypotheses that explain the observation are used to update the agent’s belief $c^*$, producing an explicit partner model $\mathcal{M}_{c^*}(\cdot)$. The model is then used by the agent to compute a joint plan $\rho^*$ for the new task. Whilst the agent computes a joint plan, it only executes its own actions. After the new task is completed, its joint trajectory becomes new evidence, and the inference-planning loop repeats.}
    \label{fig:cecm-overview}
\end{figure}

Inferring such hidden capabilities from interaction is challenging because a hypothesis can only be evaluated through the joint behaviour it induces. In Section~\ref{sec:method}, we therefore propose \textbf{C}apability \textbf{E}stimation via \textbf{C}ontextual \textbf{M}odels (\textbf{CE-CM}, overview in Figure~\ref{fig:cecm-overview}). This approximate Bayesian method infers task-invariant capability vectors through a simulate-and-compare approach: it samples candidate hypotheses, generates joint plans under them, and matches the resulting trajectories against observations. The inferred estimate then induces a capability-conditioned model for planning future tasks with the same partner. CE-CM requires no population pre-training and can refine its estimate online from a small number of interactions. However, in realistic human--AI teaming, human partners rarely act like optimal planners. Because human behaviour often involves diverse or suboptimal strategies, comparison to a single planned trajectory becomes brittle even when the capability representation is appropriate. We therefore introduce \textbf{CE-CM-Div}, an extension that evaluates each capability hypothesis against diverse planner rollouts, improving robustness to suboptimality and behavioural variability.

To evaluate our capability-based framework, we deploy it across two distinct collaborative domains: a symbolic household setting modelled in PDDL (\texttt{TidyUP}) and a simulator with under-specified rewards (\texttt{Overcooked}). In \texttt{TidyUP}, we show that CE-CM rapidly recovers representative capability estimates from only a few tasks, improves feasibility-aware coordination, and adapts when partner capabilities change over time. In \texttt{Overcooked}, we further test CE-CM in a dynamic coordination domain, where the results reveal both the promise of capability-based modelling and the limitations of assuming optimal partner behaviour. To bridge this gap, we collected 225 human \texttt{Overcooked} trajectories from 15 participants, showing that human behaviour is often not well captured by a single planned trajectory. These results motivate CE-CM-Div and show that accounting for diverse plausible behaviours substantially improves capability inference. Together, they position capability estimation as an interpretable, task-agnostic approach to adaptation in ad-hoc teamwork, while clarifying when diversity-aware inference becomes necessary. Overall, we make the following contributions in this paper:

\begin{itemize}
    \item \textbf{A capability-based formulation of multi-task ad-hoc teamwork:} We formalise repeated collaboration with unknown partners as joint planning with decentralised execution under hidden, task-invariant capability constraints. This yields an interpretable, reusable representation of what a partner can and cannot do.

    \item \textbf{CE-CM for online task-agnostic adaptation:} We introduce \textbf{CE-CM}, an approximate Bayesian method that infers discrete capability vectors from observed trajectories and uses them to induce capability-conditioned models for future planning. The method learns explicit partner models from only a small number of interactions and requires no population pre-training.

    \item \textbf{Empirical evidence for feasibility-aware coordination:} Across \texttt{TidyUP} and \texttt{Overcooked}, we show that CE-CM recovers hidden partner capabilities, reduces infeasible action assignments, and transfers estimates across tasks.

    \item \textbf{CE-CM-Div for behaviourally diverse partners:} To address the brittleness of single-trajectory inference, we propose \textbf{CE-CM-Div}, which evaluates capability hypotheses against sets of diverse planner rollouts rather than one optimal prediction. This improves capability estimation when multiple behaviours are plausible under the same capability constraints.
    
    \item \textbf{Human-subject evidence for diversity-aware inference:} Through a study of 15 participants in \texttt{Overcooked}, we show that human trajectories often differ substantially from planner-generated trajectories. This demonstrates that accounting for behavioural diversity is essential for robust capability inference in realistic human--AI teaming.
\end{itemize}

CE-CM uses the same approximate Bayesian simulate-and-compare loop as our preliminary work on CApability Modelling from Observations (CAMO) \cite{tisnikar2024probabilistic}: sampling candidate capability hypotheses, simulating them, and scoring them against observed behaviour. The setting and contribution are otherwise different. CAMO inferred continuous physical parameters from passive observations in a collaborative manipulation setting; here, an ego agent participates in repeated ad-hoc teamwork tasks while inferring discrete, task-invariant partner capabilities for future planning. Thus, this work reuses CAMO's inference principle, but extends it for online adaptation, decentralised execution, and behavioural ambiguity in multi-task human--AI teaming.

\section{Related Work} \label{sec:rw}

\subsection{Collaborating with Diverse Partners}

Most approaches that aim to create robust agents capable of cooperating with diverse partners have been developed by the AHT community, although some other fields such as zero-shot coordination (ZSC) \cite{hu2020other} (also known as convention learning \cite{shih2021Critical}) share a similar problem formulation. ZSC approaches explicitly assume a known reward function that is identical for both agents, while this might not generally be the case in AHT settings \cite{mirsky2022Survey}. Nevertheless, both fields have converged to similar solution approaches and we therefore review them together.

The key aim of the methods is to create an agent's ego policy (i.e., a mapping between joint states and the agent's individual actions) that is robust to different partners. To do so, most methods utilise population-based training, which builds on the popular self-play framework \cite{heinrich2016deep}, where the agent's robustness to variation is achieved by continuously playing the game against either itself, or its older policy checkpoints \cite{strouse2021collaborating}. 

Most research has therefore focussed on the challenge of developing populations that are diverse with respect to some metrics, such as partner trajectory diversity \cite{lupu2021trajectory, szot2023adaptive}, or partner policy diversity \cite{xing2021Learning, sarkar2023diverse}. Other methods induce population diversity by modelling additional latent rewards \cite{yu2023learning}, or introduce learned combinations of policies to further enhance diversity without increasing the population size \cite{lou2023pecan}. All above methods utilise model-free, decentralised reinforcement learning algorithms to train the ego policy as a best response to all partners in the population. 

Another approach is to build a set of policies, or a policy with latent conditioning, rather than a single ego policy. The potential partner distribution is either learned from demonstration by clustering of strategies \cite{zhao2022Coordination}, handcrafted partner models \cite{gorur2019Anticipatory, barrett2015Cooperating}, or by varying partners' reward functions \cite{li2021Individualized, ghosh2020Deployment}. The correct policy at runtime is then selected by a Bayesian inference algorithm that selects the policy from the set of policies that is considered to be the best response to the partner's behaviour. Similarly, other methods utilise a monolithic policy, but use latent conditioning to identify the partner's latent parameters at runtime \cite{ghosh2020Deployment, wang2022co}.

While above methods produce agents that are capable of adapting to diverse partners, they focus on training the policy of the ego agent rather than learning an explicit, reusable model of the current partner’s capabilities. Moreover, these methods are typically developed for settings in which the cooperative task is fixed between training and deployment, and the learned policies are not intended to transfer across tasks. They also require either a generative model of the partner population, trajectory demonstrations to induce a latent space over partners, or that the partner policies are available a priori for the ego agent to train with. In contrast, CE-CM does not require demonstrations from all potential partners, or any pre-computation of policies or populations to be able to adapt to the current partner. Furthermore, CE-CM is able to learn from any task, and learns partner-specific, but task-independent parameters, allowing it to collaborate with the same partner on any future task, not just the ones it observed so far.

\subsection{Partner Behaviour Modelling}

Partner behaviour modelling has been identified as a crucial component for successful teamwork in settings where artificial (embodied) agents team up with other agents \cite{albrecht2018Autonomous, carroll2019Utility}, or humans \cite{hiatt2017human}. The partner behaviour modelling literature has produced a very diverse set of approaches, and spans work in ad-hoc teamwork, human--agent teaming, human--robot collaboration, and others.

One of the simplest techniques to model the partner's behaviour is to predict their action based on the intelligent agent's ego policy \cite{raileanu2018modeling}. However, this method makes an explicit assumption that both agents have the same capabilities and follow the same policy (and conventions) in the task. Instead, methods within AHT literature perform model selection amongst a pool of partner models that are (partially) parametrised by their type \cite{albrecht2018Autonomous, albrecht2016Belief}. The type space specifies a set of possible partners, and by knowing a partner's type, the intelligent agent can instantiate the partner's policy. The methods use these policies to query the partner's future actions while planning the intelligent agent's course of action \cite{barrett2017Making, albrecht2019Reasoning, shafipouryourdshahi2022Online}. These methods require that the type space is known and explicit in the sense that each type in the type space induces a policy, and the policies are assumed to be known and given to the methods by the designers.

Other methods forgo the specification of a type space and instead learn the partner's representation directly from observation through repeated interaction. For example, the partner's policy is iteratively approximated by using behaviour cloning \cite{czechowski2020Decentralized}, or by biasing the planner's exploration towards previously observed joint actions \cite{wu2011online}. However, these methods require repeated interaction in a single task in order to learn these representations. While they achieve individual adaptation, they do not learn partner models that can be reused in different tasks.

Partner behaviour modelling has also been utilised in the context of goal recognition \cite{leon2022intuitive,ribeiro2021Helping,ribeiro2024HOTSPOT, huang2016Anticipatory, laidlaw2025assistancezero}, where goal-conditioned partner models can explain the partner's behaviour and allow the intelligent agent to be able to predict their future actions and adapt its own behaviour in return. The inference over possible partner goals is done using pre-computed goal-conditioned partner policies \cite{ribeiro2021Helping, ribeiro2024HOTSPOT}, or with directly querying models learned from data \cite{huang2016Anticipatory, laidlaw2025assistancezero}. These methods therefore adapt by understanding the task at hand with their partner, and then pursuing complementary actions based on this information. They do not explicitly learn partner-specific properties that can be carried across tasks.

To this end, many methods aim to learn the partner's preferences in a collaborative task. Preferences are encoded as rewards \cite{nikolaidis2015Efficient, bestick2018Learning, zhao2023learning, trivedi2018Inverse, nikolaidis2013Humanrobot,  nikolaidis2015Improved}, action costs \cite{narcomey2024Learning, canal2019Adapting}, or are simply accounted for by generating diverse plans \cite{nguyen2012Generating}.They specify the partner's preferred course of action in a given task, and can be used to predict their behaviour by performing joint planning for the team. They are learned either from passive observations \cite{trivedi2018Inverse}, active feedback queries \cite{zhao2023learning, narcomey2024Learning, bestick2018Learning}, or by swapping the agents' roles in the task \cite{nikolaidis2013Humanrobot,  nikolaidis2015Improved}. However, while preferences could in theory capture the partner's potential diversity and different capability, most learned preferences are specific to the task in which they are learned, and thus have limited transferability. Preferences therefore help explain which feasible behaviour a partner may choose within a task, whereas capabilities determine which behaviours are feasible in the first place.

Estimation of partner behaviour models is also formulated as a parameter estimation problem in centralised task allocation and scheduling approaches. The parameters that denote some behavioural descriptor of the partner such as their task proficiency \cite{emam2020Adaptive, ali2022Heterogeneous, liu2022Coordinating}, levels of attentiveness and fatigue \cite{izquierdo2022improved}, or capabilities \cite{zhang2020RealTime, fu2023Robust}. The parameter estimates are learned through direct feedback on successful execution of the allocated task or team strategy. However, in these settings, the team's actions are determined by a centralised planner or scheduler, and therefore they violate the assumption made that the partners are uncontrollable.

The parameters that govern behaviour can also be latent. Previous methods have considered learning latent decision parameters from observations \cite{orlov2022factorial} or from active collaboration \cite{unhelkar2019semi}. However, these models again rely on previously collected data that they can utilise in (semi-)supervised learning settings to explicitly learn a link function between the latent parameters and the individual actions of the partner.

While the modelling landscape is very diverse, the majority of the works focus on learning models that are specific to the context of the task at hand. Little attention, if any, has been given to learning partner models that are transferable across many collaborative tasks with the same individual. This is exactly the focus of CE-CM, as by learning a portable partner model based on their capabilities, the model is specific to the individual partner, but independent of the tasks. Because of this, it can be easily learned, refined, and reused across different tasks, and allows for effective behaviour modelling, prediction, and adaptation.

\section{Problem Formulation} \label{sec:problem}

We now describe the problem of capability inference and adaptation from observations in joint tasks. Consider two agents: $Ag_1$, who is a controllable autonomous agent, and $Ag_2$, who is an uncontrollable agent (or human), that is collaborating with $Ag_1$. The two agents collaborate in an environment to achieve a known shared goal by executing a sequence of actions, maximising some notion of reward. Both agents are characterised by capabilities that remain unchanged over the sequence of tasks. We assume that $Ag_2$ knows the true capabilities of both agents, but $Ag_2$'s true capabilities are hidden from $Ag_1$.  In order to successfully adapt, $Ag_1$ must therefore estimate $Ag_2$'s capabilities and use the estimate to plan its own contributions.

\subsection{Modelling Framework}

To model this setting, we extend the contextual MDP framework of Hallak et al.~\cite{hallak2015contextual} by defining a finite, goal-conditioned\footnote{We use goal and task interchangeably throughout this paper, and assume that any task in a goal-conditioned MMDP can be described by its goal state.}, contextual Multi-agent Markov Decision Process (CMMDP):

\begin{equation*}
    \mathcal{M}_c(g) = \langle S, A, T(s',a,s |c) ,R(s,a|g), G, C, \gamma \rangle,
\end{equation*} %\label{eq:cmdp}
where:
\begin{itemize}
    \item $S$ is a finite set of discrete joint states (we assume a global shared state between both agents);
    \item $A=A_{Ag_1} \times A_{Ag_2}$ is a finite set of joint actions, composed of individual actions $a=\{a_{Ag_1}, a_{Ag_2} \}$;
    \item $T: S \times A \times S \times C \rightarrow \{0,1\}$ is the context-dependent transition function, which we assume to be deterministic in this paper;
    \item $R(s,a|g)$ is the goal-conditioned reward function, which encodes the goal and other order-inducing information such as the agents' preferences over joint actions. The reward function is assumed to be known to both agents;
    \item  $g \in G$ represents the goal that is shared and known to both agents in advance, and is an element of all possible goals in the domain $G \subseteq S$; and
    \item $\gamma \rightarrow (0,1]$ is the discount factor. 
\end{itemize}    

We model $Ag_2$'s capabilities as $D$-dimensional vectors of binary variables $c \rightarrow \{0,1\}^D$. Each such vector induces a specific instance of $\mathcal{M}_c(g)$, and thus restricts the space of feasible joint plans. The set of context vectors $C$ is therefore the space of $Ag_2$'s possible capability vectors over which $Ag_1$ reasons.

\subsection{Capabilities as Transitions}

The capabilities modulate $\mathcal{M}_c(g)$ by restricting feasibility of $Ag_2$'s actions through the transition function $T$. Each capability, which can have semantic meaning, can influence one or more transitions, and a single transition can be influenced by more than one capability. For a transition $s \rightarrow s'$ to occur given a joint action $a=\{a_{Ag_1}, a_{Ag_2} \}$, there is a set of \emph{enabling capabilities} that must all be present for the transition to change the state. More formally, let $c' \subseteq c$ be the (possibly empty) subset of capabilities affecting the transition $\langle s, a, s' \rangle$. The effect of the enabling capabilities is as follows:

\begin{align*}
&\exists c_i \in c' \mid c_i = 0 \implies T(s', a, s \mid c) =
\begin{cases}
1 & \text{if } s' = s, \\
0 & \text{otherwise.}
\end{cases} \\
\end{align*}

We model missing capabilities as self-loops rather than removing actions from the action set, because capability requirements depend on the state context (e.g., object properties or locations), not only on the action itself. An action may therefore be executable in one state but infeasible in another.

\subsection{Inference and Planning Objectives}

Given that capabilities determine the induced CMMDP, $Ag_1$ must infer $c$ from observations in order to explain past behaviour and align its representation of $Ag_2$'s capabilities to the partner. Estimating $c$ allows $Ag_1$ to instantiate the correct transition model $T(\cdot | c)$ and thus reason about feasible joint behaviour in future tasks. To do so, $Ag_1$ seeks to align its model $\mathcal{M}_c(g)$ by finding the capabilities vector $c$ such that the predicted joint trajectories, produced by $Ag_1$'s planning process, match the team's executed trajectories. $Ag_1$ has access to observed joint state trajectories in the environment $\tau_{obs} = \langle s_i \rangle_{i=1}^H$,  as well as the known reward function. 

We assume that the executed trajectories are aligned to $Ag_2$’s true capability-constrained model. In particular, when $Ag_1$’s intended action conflicts with the joint plan induced by $Ag_2$’s true capabilities, $Ag_2$ is assumed to have the ability to correct $Ag_1$ and enforce the aligned action instead. As a result, the trajectories observed by $Ag_1$ are assumed to be feasible under the true CMMDP and they reflect coordination under $Ag_2$’s joint plan. For simplicity, we do not model these alignment actions explicitly in the CMMDP, unlike communicative teamwork formulations such as that of Pynadath and Tambe \cite{pynadath2002communicative}.

More formally, $Ag_1$ is looking to learn $c^*$, the best estimate of $c$, that minimizes the following expression: 

\begin{equation} \label{eq:scamo-objective}
    c^*=argmin_{\hat{c} \in C} \sum_{g \in G_{seen}} L \big ( \tau_{\hat{c},g}, \tau_{obs, g} \big ),
\end{equation}   
across all previous seen goals $G_{seen} \subset G$. Here, $L$ is a similarity metric, measuring the difference between the joint trajectory planned by $Ag_1$ under the current capability vector estimate $\hat{c}$, $\tau_{\hat{c},g}$, and observed joint trajectory $\tau_{obs}$. We discuss the choice of similarity metrics in Section \ref{sec:method} and~\ref{app:methods}.

Once $Ag_1$ estimates $c^*$, it can induce the corresponding task model $\mathcal{M}_{c^*}$ and compute a joint plan. Joint planning is therefore used to reason about coordination, not to control $Ag_2$, as $Ag_1$ computes a team-level plan under $c^*$, but executes only its own actions. More formally, the objective of $Ag_1$ is to find a \textit{joint} plan ${\rho = \langle a_1,a_2,...,a_H \rangle}$ that maximises the reward whilst achieving $g$, by appropriately allocating the actions to each agent. $Ag_1$ is optimising the following objective in an MMDP induced by the context $c^*$, $M_{c^*} =\langle S,A,T,R,\gamma \rangle$:

\begin{equation*}
    \rho^*(g) = \argmax_{\rho} \sum_{t=1}^{\infty} \gamma^tR(s_t,a_t|g).
\end{equation*} 

$Ag_1$ therefore needs to find a joint plan $\rho^*(g)$ that solves the task while anticipating $Ag_2$'s contributions under the estimated capability model. The resulting plan is joint in the sense that it allocates actions to both agents, but execution remains decentralised: $Ag_1$ only carries out its own actions while using the plan to anticipate $Ag_2$’s role.

In summary, we formulate adaptation as the problem of inferring a hidden partner capability vector from observed joint trajectories and using its estimate to induce a partner model for future coordination. This creates a continual estimation and planning loop, as each newly observed task provides additional evidence about the partner’s capabilities, and the updated estimate is then used to plan subsequent interactions.

\section{Methods} \label{sec:method}

In this section, we present Capability Estimation via Contextual Models (CE-CM), our method for inferring partner capabilities from observations of joint trajectories and using those estimates to support coordination in future tasks. CE-CM implements the estimation-planning loop described in Section~\ref{sec:problem}. Given observed trajectories, the agent infers a capability vector that explains past observations, and uses this estimate to induce a model for planning subsequent interactions. 

At a high level, CE-CM operates iteratively over tasks (Figure~\ref{fig:cecm-overview}, Algorithm~\ref{alg:CE-CM}). In the \emph{inference step}, candidate capability vectors are evaluated by simulating the joint trajectory they induce and comparing them to observations. Hypotheses that are consistent with the observed trajectory are retained, forming an empirical approximation of the posterior. In the \emph{planning step}, the agent uses the estimate to compute a joint plan for the next task, which is executed in a decentralised manner.

This formulation makes use of the same underlying capability model both to explain observed behaviour and to plan future actions. However, when multiple behaviours are compatible with the same capability vector, comparing observations to a single simulated trajectory can be brittle. To address this, we introduce CE-CM-Div, an extension that evaluates each capability hypothesis against a diverse set of simulated trajectories rather than a single rollout, improving robustness to behavioural variability.

\begin{algorithm}[h]
\footnotesize
\caption{Capability Estimation via Contextual Models (CE-CM)}
\label{alg:CE-CM}
\KwIn{Sequence of tasks $g_1, g_2, \dots$, similarity threshold $\varepsilon$, sample budget $N$}
\KwOut{Capability estimates $\hat{c}^*$ and plans $\rho^*(g)$}

Initialise sample set $Q_0 \gets \emptyset$\;

\For{each task $g_n$}{
    Observe trajectory $\tau_{obs, g_n}$\;
    
    $Q_n \gets Q_{n-1}$\;
    
    \For{$i = 1$ to $N$}{
        $c \sim \textsc{Prior}(c)$\;
        $\tau_{c, g_n} \gets \textsc{JointPlanner}(c, g_n)$\;
        $L \gets \textsc{Similarity}(\tau_{obs, g_n}, \tau_{c, g_n})$\;
        
        \If{$L < \varepsilon$}{
            $Q_n \gets Q_n \cup c$\;
        }
    }
    
    $\hat{c}^* \gets \textsc{Estimate}(Q_n)$\;
    
    \If{next task $g_{n+1}$ exists}{
        $\rho^*(g_{n+1}) \gets \textsc{JointPlanner}(\hat{c}^*, g_{n+1})$\;
    }
}
\end{algorithm}

\subsection{Updating the Capability Belief with Approximate Inference}

To update the intelligent agent's beliefs over partner's capabilities, the agent seeks to infer a posterior distribution $P(c|\tau_{obs})$  over the partner's capability vector given the observed joint trajectory. In principle, the agent could update the posterior using Bayesian updating: $P(c|\tau) \propto \mathcal{L}(\tau|c) \cdot P(c)$, where the posterior is proportional to the likelihood of the observed trajectory under a capability hypothesis and a prior over capabilities. However, evaluating the likelihood $\mathcal{L}(\tau|c)$ is intractable in our setting, as it requires solving the forward planning problem under each candidate capability vector. The behaviour induced by a given $c$ depends on both the reward structure and the capability-conditioned transition function, making direct likelihood computation impractical.

We therefore approximate this likelihood using simulation-based inference, specifically Approximate Bayesian Computation (ABC) \cite{beaumont2019approximate}. The key idea is to evaluate capability hypotheses by generating trajectories under each candidate vector $c$ and comparing them to the observed trajectory, accepting those that are sufficiently similar.

Formally, this corresponds to sampling from an approximate posterior (lines 4–9 in Algorithm~\ref{alg:CE-CM}):
\[
P_{\epsilon}(c \mid \tau_{obs}) = \{c : L(\tau_{c}, \tau_{obs}) < \epsilon\},
\]
where $\epsilon > 0$ is a similarity threshold, and $L$ is a distance metric between trajectories. The choice of $L$ depends on the state representation: we use Jaccard distance for propositional (PDDL) states, and cosine similarity for real-valued state vectors.

Let 
\[
Q_n = \{c : L(\tau_{c, g_n}, \tau_{obs, g_n}) < \epsilon\}
\]
be the multi-set of capability vectors that produce trajectories sufficiently similar to the observation in task $g_n$. Intuitively, $Q_n$ contains capability hypotheses that are consistent with the observed behaviour in task $g_n$. The union of accepted samples across tasks, $\bigcup_{j=1}^{n} Q_j$, aggregates evidence over time and acts as an empirical approximation of samples from the posterior. CE-CM uses this accumulated sample set to compute a point estimate $\hat{c}^*$ of the partner’s capabilities (line 11 in Algorithm~\ref{alg:CE-CM}).

To obtain a parametric estimate, we approximate the posterior as a multivariate Bernoulli distribution. Each capability $c_i$ is treated as an independent Bernoulli random variable with parameter $\theta_i \equiv P(c_i = 1)$, so that the belief at iteration $n$ is represented as a vector $\beta^n(c) = [\theta_1, \dots, \theta_D]^\top$. The parameters in $\beta$ are updated by computing the empirical mean over accepted samples:
\[
\beta^{n+1}(c_i) = \frac{\left|\bigcup_{j=1}^{n} Q_j^{+}(c_i)\right|}{\left|\bigcup_{j=1}^{n} Q_j\right|},
\]
where $Q_j^{+}(c_i) = \{\hat{c} \in Q_j \mid \hat{c}_i = 1\}$ are all accepted samples in which the capability $\hat{c}_i$ was present. Finally, the MAP estimate is obtained by thresholding:
\[
\hat{c}_i^* =
\begin{cases}
1 & \text{if } \beta^{n+1}(c_i) > \psi, \\
0 & \text{otherwise.}
\end{cases}
\]
The resulting estimate $\hat{c}^*$ can be interpreted as an approximate solution to Equation~\ref{eq:scamo-objective}, as it selects the capability vector that best explains the observed trajectories under the ABC approximation, and is used to induce a capability-conditioned model for planning future interactions.

\subsection{Planning with Capability Estimates}

Once CE-CM obtains an estimate $\hat{c}^*$ of the partner’s capabilities, it uses this estimate to induce a capability-conditioned model $\mathcal{M}_{\hat{c}^*}(g)$ and compute a joint plan for the next task (line 13 in Algorithm~\ref{alg:CE-CM}).

Planning is performed using a joint planner over the induced CMMDP, assigning actions to both agents while respecting the capability constraints. As discussed in Section~\ref{sec:problem}, execution remains decentralised: $Ag_1$ carries out only its own actions, using the joint plan to anticipate the partner’s behaviour. The specific planning implementation depends on the domain representation and is described in~\ref{app:methods}.

\subsection{CE-CM-Div: Extending CE-CM to Suboptimal and Diverse Trajectories}

CE-CM implicitly assumes that each capability vector induces a single joint trajectory that the partner will follow. For this assumption to hold, the agents' plans must be optimal with respect to their capabilities, and their preferences must be fully specified to induce strong orderings on behaviours. In practice, agents may act suboptimally, follow unmodelled preferences, or choose among multiple behaviours with equal returns, giving rise to multiple plausible trajectories under the same capabilities. In such cases, comparing observations to a single simulated trajectory can cause CE-CM to incorrectly reject valid capability hypotheses.

To address this, we introduce CE-CM-Div, an extension that modifies the likelihood approximation in CE-CM. Instead of generating a single trajectory for each capability hypothesis, CE-CM-Div evaluates each hypothesis against a set of diverse trajectories $\Delta$ that capture different feasible behaviours under that capability vector. For each candidate capability vector $\hat{c}$, the planner produces a set of trajectories $\Delta_{\hat{c},g}$ for the observed goal $g$, and the hypothesis is accepted when at least one trajectory in this set falls within the ABC tolerance under the trajectory-distance metric $L$, i.e., when $\min_{\tau \in \Delta_{\hat{c},g}} L(\tau, \tau_{obs,g}) \leq \epsilon$. This prevents rejecting a capability vector due to the single rollout produced by the planner differs from the human's chosen behaviour. We consider such rejections as false negatives, because the capability hypothesis is feasible and compatible with the observation, but is incorrectly discarded due to unmodelled behavioural diversity.

This modification replaces single-trajectory comparison in CE-CM with a set-based comparison, while leaving the rest of the inference and planning loop unchanged. In practice, the diverse trajectory set $\Delta$ can be generated using any planner capable of producing multiple distinct solutions. In our implementation, we use a diversity-aware variant of Monte Carlo Tree Search (MCTS) to generate a set of trajectories that differ in their visited states, following prior work on diverse planning \cite{benke2023diverse, srivastava2007domain}. Full implementation details are provided in~\ref{app:ce-cm-div}.

\section{Experiments} \label{sec:exp}

We evaluate CE-CM along three core hypotheses:

\begin{itemize}
    \item \textbf{H1 (Capability inference):} CE-CM can recover a partner’s latent capability vector from interaction across tasks.
    \item \textbf{H2 (Coordination impact):} Improved capability estimates lead to better coordination, reflected in reduced misallocation of actions and increased agreement between agents.
    \item \textbf{H3 (Robustness to behavioural variability):} When observed behaviour is not well explained by a single trajectory, CE-CM-Div improves capability inference over CE-CM by accounting for multiple plausible behaviours.
\end{itemize}

We test these hypotheses in two collaborative domains with complementary properties\footnote{All of our code is available at \url{https://github.com/Ptisni/CE-CM}}. \texttt{TidyUP} is a PDDL 2.1 \cite{fox2003pddl2} domain with fully specified rewards and strong ordering over joint plans, allowing us to evaluate capability inference and its downstream impact in a well-defined setting. \texttt{Overcooked} is a simulator-based coordination domain in which multiple joint behaviours can achieve similar outcomes, making it suitable for evaluating the limits of capability-based reasoning and the effects of behavioural variability. In addition to simulated partners, we collect human gameplay data in \texttt{Overcooked} to evaluate performance under realistic behaviour.

To evaluate these hypotheses, we use the following metrics:

\begin{itemize}
    \item \textbf{Capability inference accuracy}: Hamming distance between the inferred and ground-truth capability vectors, measuring how well the agent recovers the partner’s capabilities.
    \item \textbf{Coordination quality}: (i) the ratio of corrections, capturing how often the agent proposes actions that conflict with the partner’s behaviour, and (ii) plan overlap, measuring agreement in action allocation between the agents.
    \item \textbf{Assigned action feasibility}: the proportion of unproductive (infeasible) actions assigned to the partner, and the proportion of feasible actions in open-loop planning on unseen tasks.
\end{itemize}

The remainder of this section is organised as follows. In Section~\ref{sec:exp-tidyup}, we evaluate CE-CM in \texttt{TidyUP}, focusing on capability inference and its impact on coordination in a well-specified setting. In Section~\ref{sec:exp-overcooked}, we analyse CE-CM in \texttt{Overcooked} with simulated partners, highlighting both its strengths and its limitations when multiple behaviours are compatible with the same capabilities. Finally, in Section~\ref{sec:exp-human}, we evaluate CE-CM-Div on human data and show that accounting for behavioural diversity substantially improves capability inference in realistic settings.

\subsection{TidyUP} \label{sec:exp-tidyup}

\begin{figure}[h!]
    \centering
    \includegraphics[width=0.7\linewidth]{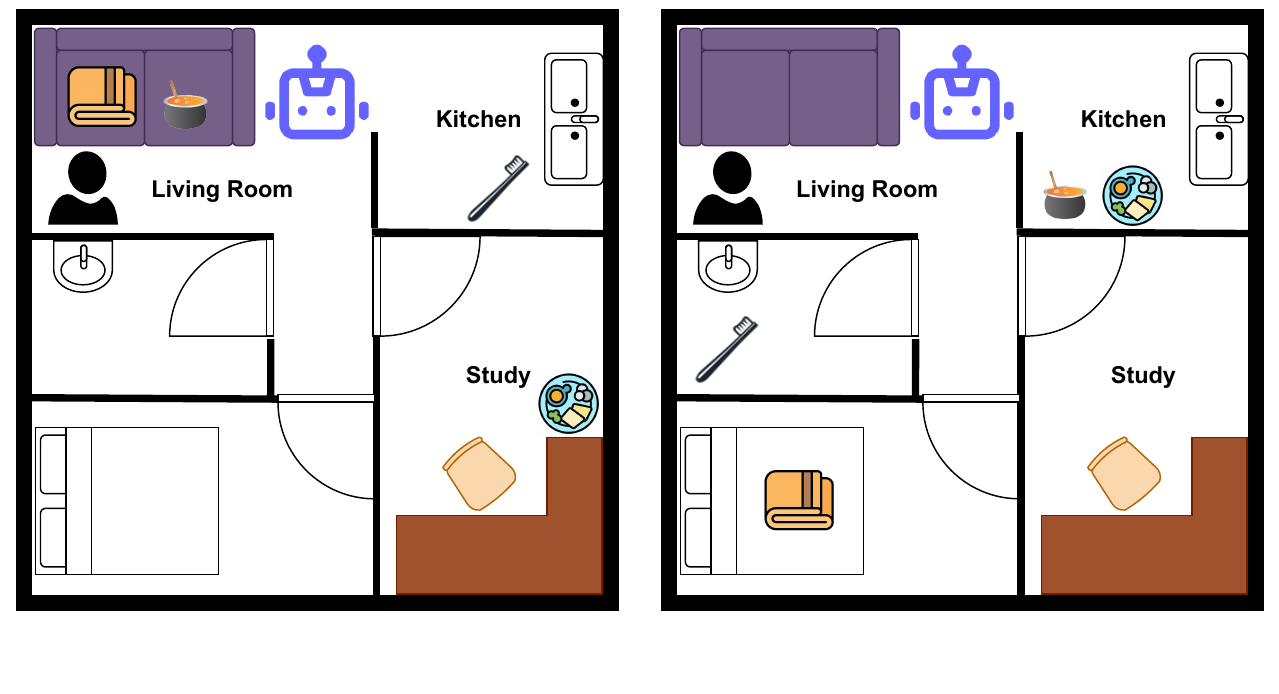}
    \caption{\texttt{TidyUP} domain. Left: one of the possible starting states, where the blanket and a dirty bowl are on the couch, the dirty dish is in the study, and the toothbrush is in the kitchen. Right: possible goal state where the blanket is on the bed, the bowl and the dish are clean and in the kitchen, and the toothbrush is in the bathroom.}
    \label{fig:env-tidyup}
\end{figure}

We first evaluate CE-CM in \texttt{TidyUP}, a PDDL-based household domain in which a human and their robot assistant jointly tidy the apartment by cleaning and rearranging objects across several rooms (see Figure~\ref{fig:env-tidyup} for an example scenario, and~\ref{app:domains-tu} for a more detailed description). The joint state encodes object locations, object cleanliness, and agent positions. Transitions are deterministic and governed by action preconditions, which correspond directly to capabilities (e.g., an agent can only pick up a plate if \texttt{can\_pick(plate)} is true). The reward function specifies action costs, inducing a clear ordering over joint plans.

The domain contains 20 capabilities, yielding a 20-dimensional capability vector. The robot is fully capable, while the simulated human exhibits restricted capabilities. We consider two partner types:
\begin{enumerate}
    \item Low-capability (LC) human, who can move freely and manipulate objects only in the bathroom and kitchen; and
    \item High-capability (HC) human, who can perform all actions except placing objects in the bedroom and living room.
\end{enumerate}

The simulated human plans using the ground-truth CMMDP $\mathcal{M}_c$ and has full knowledge of both agents' capabilities, while the robot must infer the human’s capabilities from interaction. Both agents plan independently but execute only their own actions. To ensure that the executed trajectory reflects the true capability-constrained model, we allow the human to correct the robot’s actions when they conflict with the human’s plan, effectively enforcing alignment during execution. This mirrors the alignment assumption introduced in Section~\ref{sec:problem}.

\subsubsection{Experimental Setup and Evaluation}

Each experimental run consists of a sequence of 5 tasks sampled from the domain's goal set. After each observed task, CE-CM updates its capability estimate and uses it to plan the next task. We repeat this process over 20 independent runs. This setup directly evaluates \textbf{H1} (capability inference) and \textbf{H2} (coordination impact) as the agent accumulates experience over tasks.

We compare CE-CM against two baselines. The \emph{optimistic} baseline assumes the partner is fully capable and does not adapt over time. The \emph{pessimistic} baseline enables capabilities only after they are directly observed through successful transitions, effectively following a ``what you see is what you believe'' strategy. While this baseline can perform well in domains with explicit action preconditions, such as PDDL, it relies on knowing the mapping between transitions and capabilities, which may not generalise to other settings. All implementation details for CE-CM and the baselines are provided in~\ref{app:methods-tu}.

\subsubsection{Results}

\begin{figure}[t]
     \centering
     \begin{subfigure}{0.45\textwidth}
         \centering
         \includegraphics[width=\linewidth]{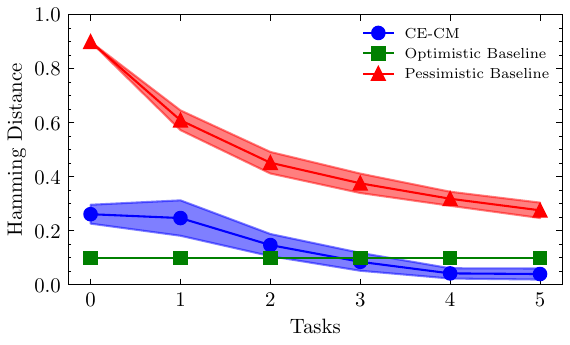}
         \caption{High-capability type}
     \end{subfigure}
     \begin{subfigure}{0.45\textwidth}
         \centering
         \includegraphics[width=\linewidth]{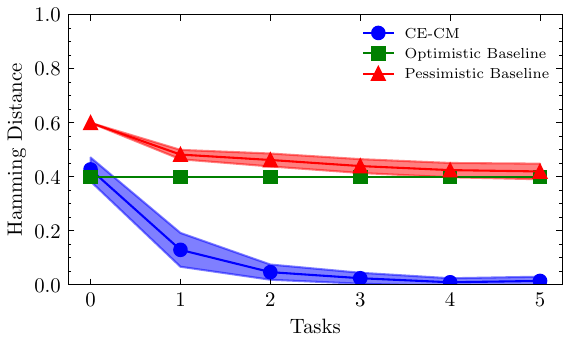}
         \caption{Low-capability type}
     \end{subfigure}
     \caption{Hamming distance between ground truth and inferred capability vectors. Solid line indicates the mean, and shaded area denotes the 95\% Confidence Interval.}
     \label{fig:hamming_distances}
\end{figure}

We first evaluate whether CE-CM recovers the partner's capabilities from interaction. Figure~\ref{fig:hamming_distances} shows that CE-CM rapidly reduces Hamming distance for both partner types, converging to near-zero error within five observed tasks. In contrast, the optimistic baseline fails to correct its initial assumptions, while the pessimistic baseline converges more slowly due to relying solely on direct observation. These results demonstrate that CE-CM can efficiently infer the partner's capability vector from a small number of collaborative tasks.

\begin{figure}[t]
     \centering
     \begin{subfigure}{0.45\textwidth}
        \centering
        \includegraphics[width=\linewidth]{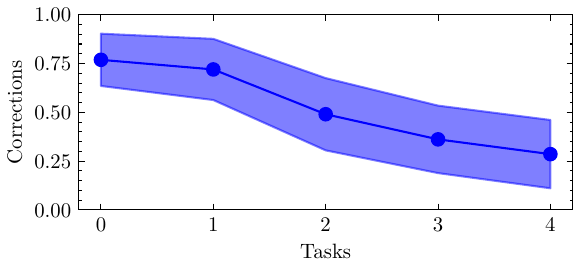}
        \caption{High-capability type}
     \end{subfigure}
     \begin{subfigure}{0.45\textwidth}
         \centering
         \includegraphics[width=\linewidth]{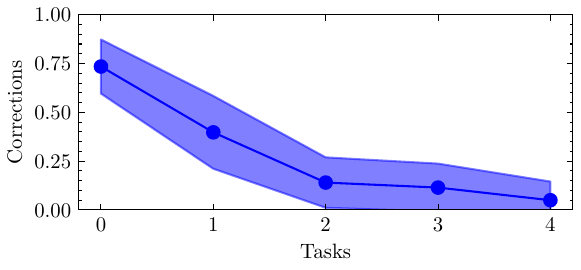}
         \caption{Low-capability type}
     \end{subfigure}
     \caption{Ratio of corrected actions across tasks. Solid line indicates the mean, and shaded area denotes the 95\% Confidence Interval.}
     \label{fig:disagreements}
\end{figure}

We next evaluate whether improved capability estimates lead to better coordination. Figure~\ref{fig:disagreements} shows that the proportion of corrected actions decreases substantially as more tasks are observed, dropping from approximately $75\%$ to below $30\%$ for HC and below $5\%$ for LC. This indicates that the agent increasingly assigns actions in a manner consistent with the partner’s capabilities.

\begin{figure}[t]
     \centering
     \begin{subfigure}{0.45\textwidth}
        \centering
        \includegraphics[width=\linewidth]{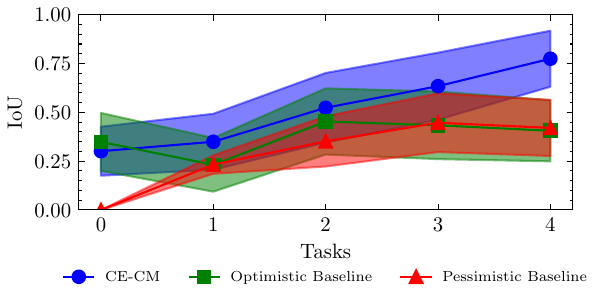}
        \caption{High-capability type}
     \end{subfigure}
     \begin{subfigure}{0.45\textwidth}
         \centering
         \includegraphics[width=\linewidth]{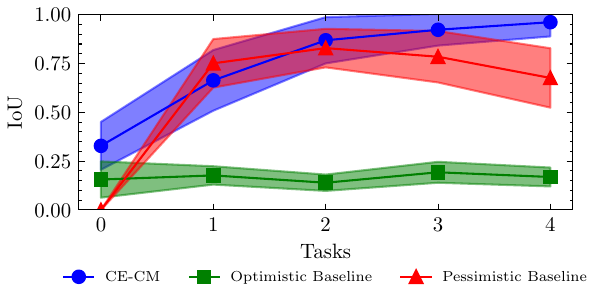}
         \caption{Low-capability type}
     \end{subfigure}
     \caption{Plan overlap (IoU) between joint plans. Solid line indicates the mean, and shaded area denotes the 95\% Confidence Interval.}
     \label{fig:iou}
\end{figure}

However, correction rates do not fully vanish. This is partly due to temporal mismatches: even when both agents assign actions to the same agent, differences in execution order may lead to corrections. To better capture agreement in task allocation, we measure plan overlap using IoU (Figure~\ref{fig:iou}). CE-CM achieves substantial improvements, reaching over $75\%$ overlap for HC and above $95\%$ for LC, outperforming both baselines. This indicates that CE-CM correctly identifies which agent should perform which actions, even when execution order varies.

In summary, we have shown that in a well-specified domain such as \texttt{TidyUP}, CE-CM accurately infers the partner’s capabilities and uses them to produce joint plans that align closely with the partner’s behaviour, supporting both \textbf{H1} and \textbf{H2}.

\subsection{Overcooked} \label{sec:exp-overcooked}

\begin{figure}[t!]
    \centering
    \includegraphics[width=0.55\textwidth]{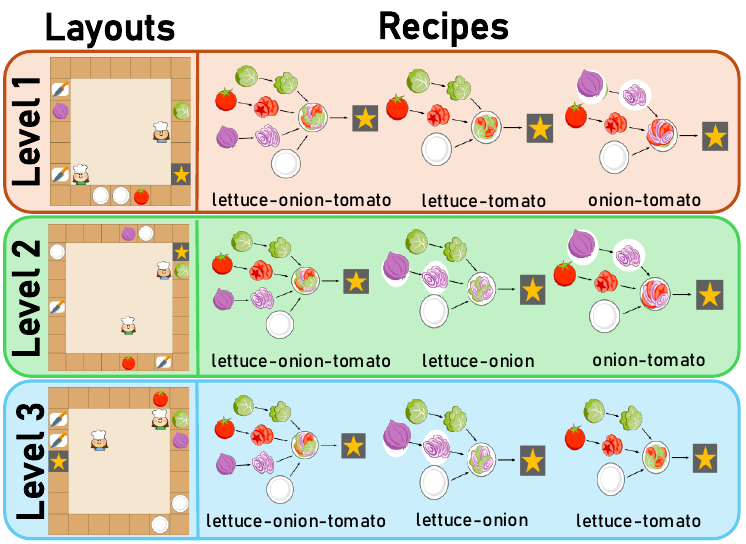}
    \caption{Overview of \texttt{Overcooked}. There are three distinct layouts, and three possible recipes, creating nine distinct tasks in total. The agents must collaborate to deliver the dish as quickly as possible, ensuring that the dish consists of the correct ingredients.}
    \label{fig:oc-overview}
\end{figure}

We next evaluate CE-CM in \texttt{Overcooked}, a common collaborative benchmark \cite{carroll2019Utility, wu2021Too} in which two agents must coordinate to prepare and deliver dishes by gathering ingredients, chopping, plating, and serving (Figure~\ref{fig:oc-overview}, see~\ref{app:domains-oc} for details). We use a macro-action variant of the environment \cite{xiao2025asynchronous}, where low-level controls are abstracted into high-level actions such as \emph{get tomato}, \emph{chop}, and \emph{deliver}.

The capability vector in this domain is 10-dimensional. The intelligent agent (blue) is fully capable, while the partner agent (green) may lack capabilities. We consider three partner types:
\begin{enumerate}
    \item Type 1: cannot handle any vegetables (cannot pick up lettuce, onion, or tomato);
    \item Type 2: cannot cut or deliver dishes;
    \item Type 3: cannot handle plates or deliver dishes.
\end{enumerate}
All partner types are controlled by the same joint planner, operating in the ground-truth CMMDP $\mathcal{M}_c(g)$.

Compared to \texttt{TidyUP}, \texttt{Overcooked} exhibits significant underspecification: many distinct joint plans can achieve similar returns. As a result, small differences in capabilities can lead to qualitatively different behaviours, and coordination becomes sensitive to conventions rather than feasibility alone. For example, multiple valid strategies exist for assembling a dish, even though the temporal structure of sub-tasks (e.g., chopping before plating) remains constrained. This makes \texttt{Overcooked} a suitable testbed for evaluating the limits of capability-based reasoning.

\subsubsection{Experimental Setup and Evaluation}

We sample 5 tasks for learning from the set of 9 layout–recipe combinations (Figure~\ref{fig:oc-overview}). After each observed task, CE-CM updates its capability estimate and uses it to plan the next task. To evaluate open-loop generalisation, we assess performance on the remaining 4 held-out tasks.

We conduct two main evaluations. First, we measure \textbf{H1} capability inference accuracy  over 20 runs per partner type using Hamming distance. Second, we evaluate \textbf{H2} downstream coordination by measuring the proportion of infeasible (unproductive) actions assigned to the partner, and the proportion of feasible actions in open-loop planning on held-out tasks.

We compare CE-CM against an optimistic baseline that assumes the partner is fully capable and does not adapt over time. We do not include the pessimistic baseline from \texttt{TidyUP}, as the mapping between individual transitions and capabilities is not explicitly defined in this domain (i.e., the simulator admits the whole capability vector at initialisation).

\subsubsection{Results}

\begin{figure}[t]
     \centering
     \begin{subfigure}{0.32\textwidth}
        \centering
        \includegraphics[width=\linewidth]{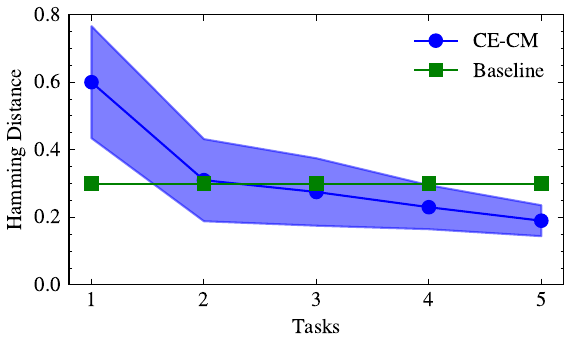}
        \caption{Type 1}
        \label{fig:hamming-1}
     \end{subfigure}
     \begin{subfigure}{0.32\textwidth}
        \centering
        \includegraphics[width=\linewidth]{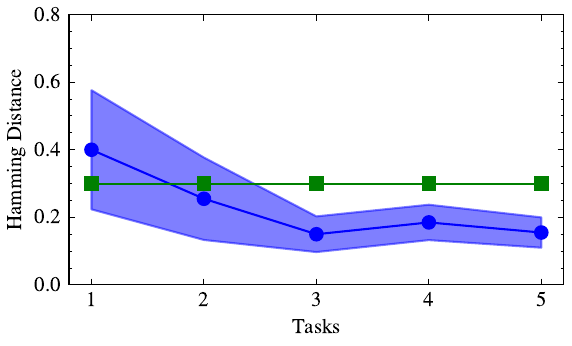}
        \caption{Type 2}
     \end{subfigure}
     \begin{subfigure}{0.32\textwidth}
         \centering
         \includegraphics[width=\linewidth]{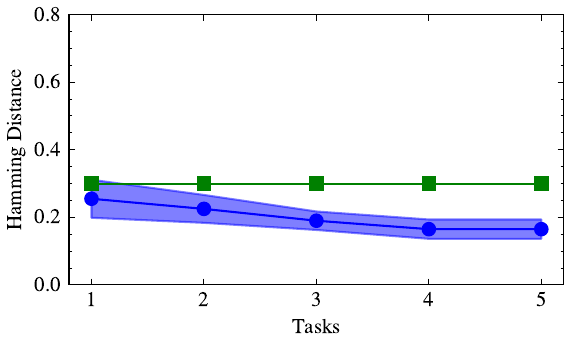}
         \caption{Type 3}
     \end{subfigure}
     \caption{Hamming distance between ground truth capabilities and the estimated capability vector. Solid line indicates the mean, and shaded area denotes the 95\% Confidence Interval.}
     \label{fig:hamming-oc}
\end{figure}

We first evaluate whether CE-CM recovers partner capabilities from repeated interaction. Figure~\ref{fig:hamming-oc} shows that CE-CM consistently reduces Hamming distance across all three partner types, reaching values between $0.18$ and $0.2$ after five tasks. Importantly, CE-CM is able to reduce error even when the exact capability vector is not sampled from the prior (e.g., Type 1, Figure~\ref{fig:hamming-1}), indicating that the method can recover a useful approximation of the underlying capability structure. In all cases, CE-CM significantly outperforms the optimistic baseline.

\begin{figure}
     \centering
     \begin{subfigure}{0.32\textwidth}
        \centering
        \includegraphics[width=\linewidth]{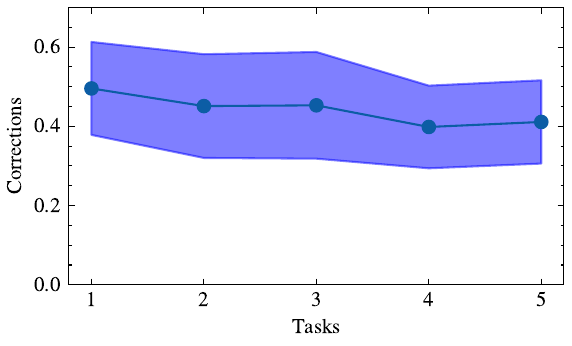}
        \caption{Type 1}
     \end{subfigure}
     \begin{subfigure}{0.32\textwidth}
        \centering
        \includegraphics[width=\linewidth]{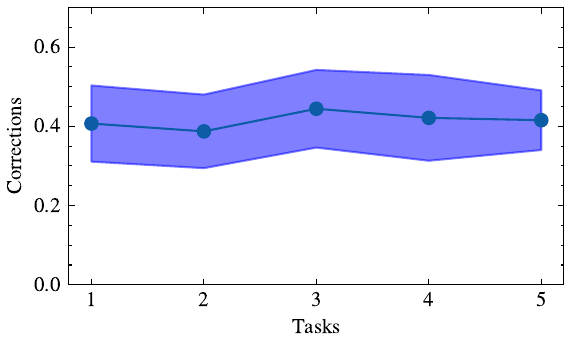}
        \caption{Type 2}
     \end{subfigure}
     \begin{subfigure}{0.32\textwidth}
         \centering
         \includegraphics[width=\linewidth]{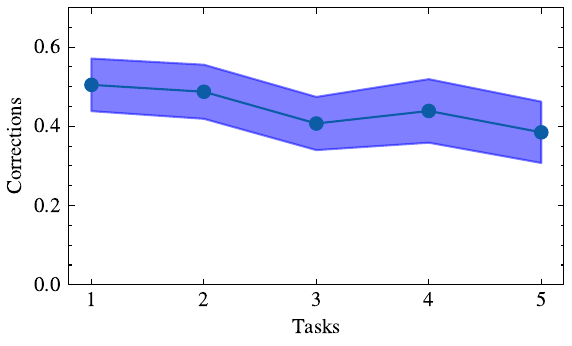}
         \caption{Type 3}
     \end{subfigure}
     \caption{Ratio of corrected actions across tasks. Solid line indicates the mean, and shaded area denotes the 95\% Confidence Interval.}
     \label{fig:corrections-oc}
\end{figure}

We next examine whether improved capability estimates translate into better coordination. Unlike in \texttt{TidyUP}, Figure~\ref{fig:corrections-oc} shows little change in the proportion of corrected actions for Types 1 and 2, with only a modest reduction for Type 3. This indicates that, despite improving capability estimates, the agent does not converge to the same joint behaviour as its partner.

This discrepancy arises from the underspecified nature of the domain. Multiple joint plans can achieve similar returns, meaning that knowing the correct capability constraints does not uniquely determine the partner’s behaviour. 

\begin{figure}
     \centering
     \begin{subfigure}{0.32\textwidth}
        \centering
        \includegraphics[width=\linewidth]{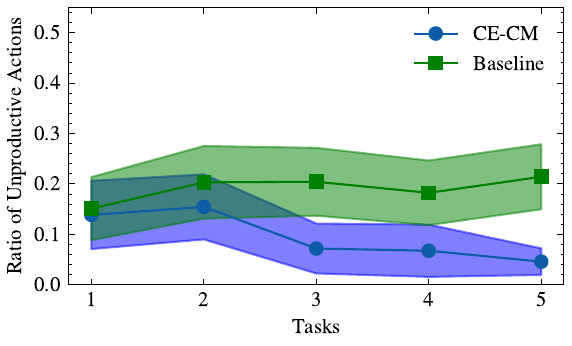}
        \caption{Type 1}
        \label{fig:illegal-1}
     \end{subfigure}
     \begin{subfigure}{0.32\textwidth}
        \centering
        \includegraphics[width=\linewidth]{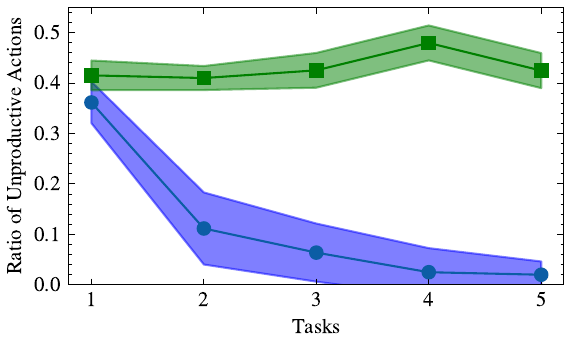}
        \caption{Type 2}
        \label{fig:illegal-2}
     \end{subfigure}
     \begin{subfigure}{0.32\textwidth}
         \centering
         \includegraphics[width=\linewidth]{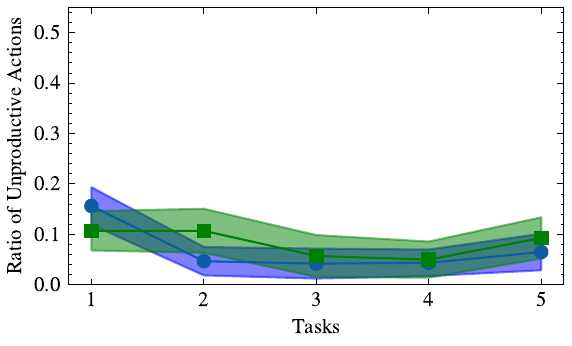}
         \caption{Type 3}
         \label{fig:illegal-3}
     \end{subfigure}
     \caption{Ratio of unproductive (infeasible) actions across tasks. Solid line indicates the mean, and shaded area denotes the 95\% Confidence Interval.}
     \label{fig:illegal-oc}
\end{figure}

We confirm this by examining the proportion of infeasible actions assigned to the partner. Figure~\ref{fig:illegal-oc} shows that CE-CM substantially reduces unproductive actions across all partner types, indicating that it correctly captures feasibility constraints even when it fails to match the partner’s exact behaviour.

\begin{figure}
     \centering
     \begin{subfigure}{0.32\textwidth}
        \centering
        \includegraphics[width=\linewidth]{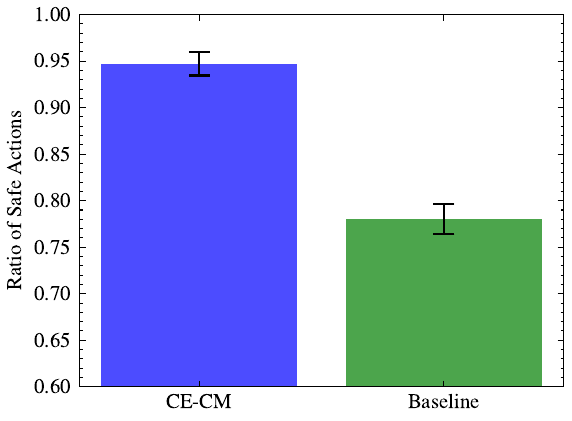}
        \caption{Type 1}
        \label{fig:feasible-1}
     \end{subfigure}
     \begin{subfigure}{0.32\textwidth}
        \centering
        \includegraphics[width=\linewidth]{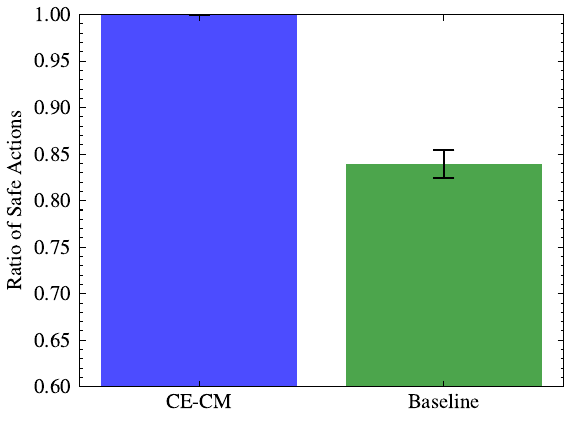}
        \caption{Type 2}
        \label{fig:feasible-2}
     \end{subfigure}
     \begin{subfigure}{0.32\textwidth}
         \centering
         \includegraphics[width=\linewidth]{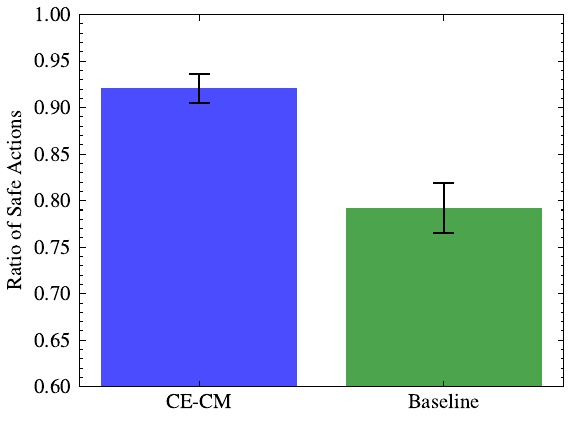}
         \caption{Type 3}
         \label{fig:feasible-3}
     \end{subfigure}
     \caption{Proportion of feasible actions in open-loop planning on held-out tasks. Solid line indicates the mean, and shaded area denotes the 95\% Confidence Interval.}
     \label{fig:feasible-oc}
\end{figure}

We further evaluate CE-CM in open-loop planning on held-out tasks. Figure~\ref{fig:feasible-oc} shows that CE-CM consistently produces plans with a high proportion of feasible actions, outperforming the optimistic baseline across all partner types. This demonstrates that CE-CM improves safety and robustness by avoiding infeasible action assignments.

In the \texttt{Overcooked} domain, CE-CM successfully recovers capability constraints (\textbf{H1}) and improves feasibility and safety of action assignment, but does not fully resolve coordination (\textbf{H2}). This highlights a key limitation: capabilities alone are insufficient to predict behaviour when multiple strategies are equally valid.

\subsection{Capability Inference from Human Players} \label{sec:exp-human}

The results in \texttt{Overcooked} highlight a key limitation of CE-CM: when multiple behaviours are consistent with the same capability vector, matching observations to a single simulated trajectory becomes unreliable. To evaluate this failure mode in a realistic setting, and to test whether CE-CM-Div addresses it, we conduct a human-subject study in the \texttt{Overcooked} domain.

Human behaviour in this setting exhibits substantial variability even within the same capability profile and task. Different players may adopt distinct strategies, conventions, or preferences, all of which can produce valid but diverse trajectories under identical capabilities. This form of behavioural ambiguity directly violates the single-trajectory assumption underlying CE-CM and provides a natural testbed for evaluating CE-CM-Div. Additional details and analysis of the dataset are provided in~\ref{app:dataset}.

\subsubsection{Experimental Setup and Evaluation}

\begin{figure}[t]
    \centering
    \includegraphics[width=0.55\linewidth]{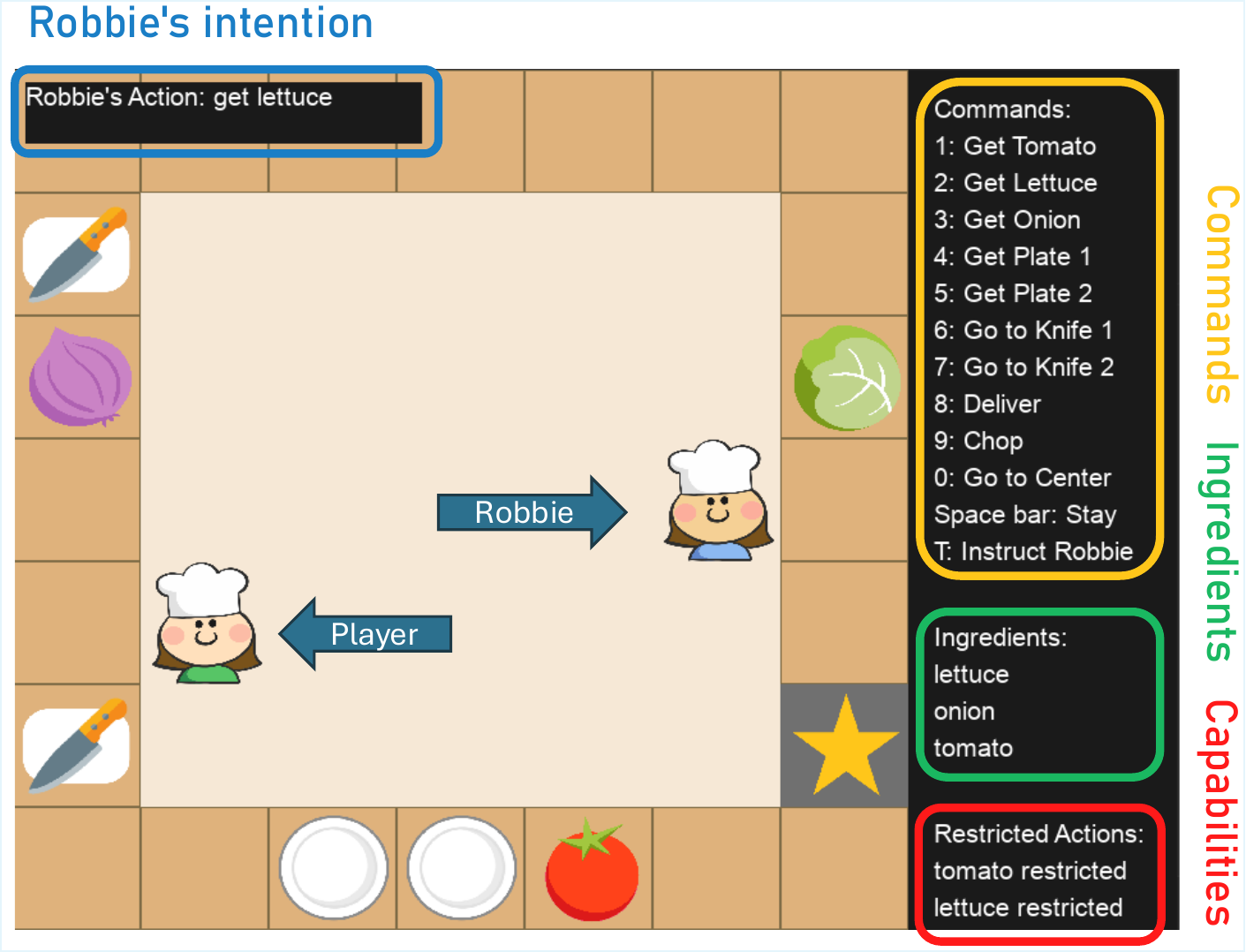}
    \caption{Overview of the \texttt{Overcooked} user interface. In the top left corner, the intelligent assistant agent indicates its next action. The available commands, the current recipe, and the restricted capabilities are displayed on the right hand side.}
    \label{fig:interface}
\end{figure}

To evaluate CE-CM and CE-CM-Div under realistic behavioural variability, we collect a dataset of human–agent interactions in the \texttt{Overcooked} domain\footnote{We obtained permission to conduct this study from King's College London Research Ethics Office (Study registration number: MRSP-24/25-50047).}. We recruited 15 participants, each of whom completed 15 episodes in collaboration with a fixed-policy assistant agent. Tasks were sampled from the same nine layout–recipe combinations used in previous experiments, with each participant interacting across all three capability types.

At each decision point, participants selected high-level actions (e.g., \emph{get ingredient}, \emph{chop}, \emph{deliver}) and were shown the assistant’s intended action, which they could correct if needed (the interface and the assistant can be seen in Figure~\ref{fig:interface}). This interaction protocol mirrors the correction-based execution model used in our simulations, ensuring consistency between experimental settings.

We record sequences of joint states along with the ground-truth capability vector associated with each trajectory, yielding a dataset of 225 trajectories (75 per capability type). Full details of data collection and participant demographics are provided in~\ref{app:data-collection}.

We evaluate both methods in an offline setting, where capability vectors are inferred directly from recorded trajectories. Our primary objective is to assess whether CE-CM-Div improves capability inference over CE-CM when learning from human behaviour.

\subsubsection{Results}

\begin{figure}[ht!]
     \centering
     \begin{subfigure}{0.32\textwidth}
        \centering
        \includegraphics[width=\linewidth]{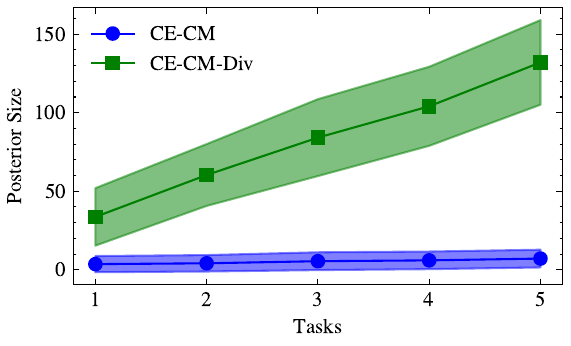}
        \caption{Type 1}
     \end{subfigure}
     \begin{subfigure}{0.32\textwidth}
        \centering
        \includegraphics[width=\linewidth]{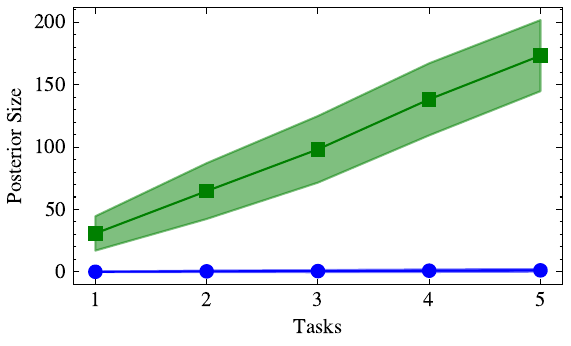}
        \caption{Type 2}
     \end{subfigure}
     \begin{subfigure}{0.32\textwidth}
        \centering
         \includegraphics[width=\linewidth]{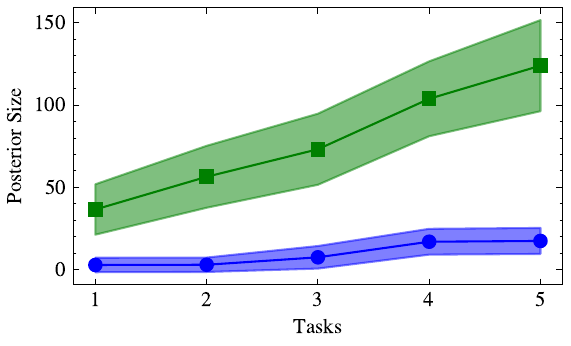}
         \caption{Type 3}
     \end{subfigure}
     \caption{Number of accepted samples in the posterior across observed tasks. Solid line indicates the mean, and shaded area denotes the 95\% Confidence Interval.}
     \label{fig:samples-diverse}
\end{figure}

We first compare CE-CM and CE-CM-Div in terms of their ability to generate consistent posterior updates. Figure~\ref{fig:samples-diverse} shows a stark difference in sample acceptance rates: CE-CM accepts very few capability samples per task (typically fewer than $10$), whereas CE-CM-Div admits substantially more (between $120$ and $175$ after five tasks). This indicates that CE-CM frequently fails to generate a trajectory sufficiently similar to human behaviour, preventing it from updating its belief. In contrast, CE-CM-Div matches observed behaviour more reliably by evaluating each hypothesis against a set of trajectories.

\begin{figure}[t]
     \centering
     \begin{subfigure}{0.32\textwidth}
        \centering
        \includegraphics[width=\linewidth]{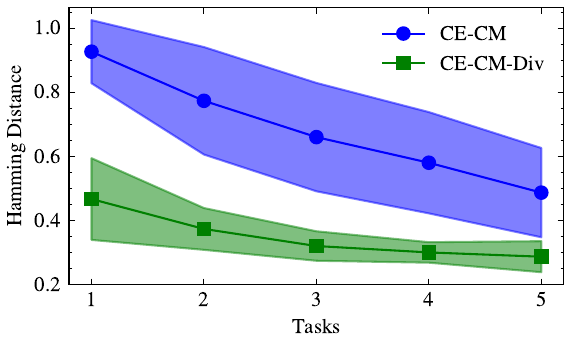}
        \caption{Type 1}
        \label{fig:distance-1}
     \end{subfigure}
     \begin{subfigure}{0.32\textwidth}
        \centering
        \includegraphics[width=\linewidth]{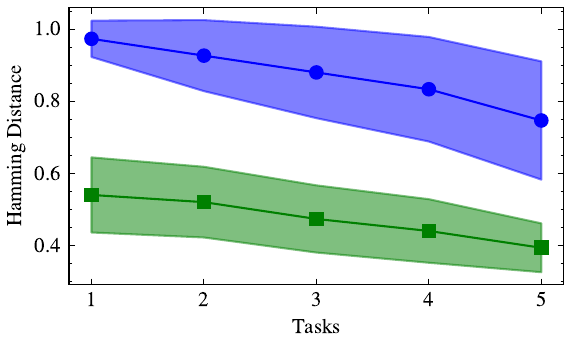}
        \caption{Type 2}
        \label{fig:distance-2}
     \end{subfigure}
     \begin{subfigure}{0.32\textwidth}
         \centering
         \includegraphics[width=\linewidth]{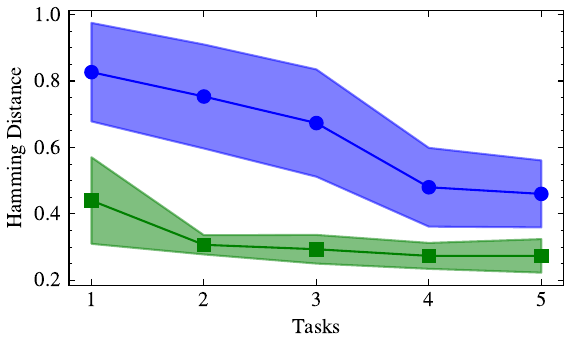}
         \caption{Type 3}
         \label{fig:distance-3}
     \end{subfigure}
     \caption{Hamming distance between ground truth and inferred capability vectors. Solid line indicates the mean, and shaded area denotes the 95\% Confidence Interval.}
     \label{fig:distances-diverse}
\end{figure}

We next evaluate capability estimation accuracy directly. Figure~\ref{fig:distances-diverse} shows that CE-CM-Div substantially outperforms CE-CM across all capability types, reducing Hamming distance to approximately $0.29$, $0.39$, and $0.27$ for Types 1, 2, and 3, respectively. In contrast, CE-CM achieves significantly higher error ($0.49$, $0.75$, and $0.46$) and frequently fails to update its estimate. When CE-CM does update, it can produce accurate estimates, but this depends on the human following a trajectory similar to the planner’s prediction. 

These results demonstrate that CE-CM-Div provides a robust solution to capability inference under behavioural variability, producing richer posterior estimates and significantly improving accuracy over CE-CM. Additional analysis of the influence of diversity parameter $\delta$ on CE-CM-Div's performance is provided in~\ref{app:diversity}.

\section{Discussion} %\label{sec:discussion}

Our experimental results show that CE-CM can reliably recover a partner’s capabilities in well-specified settings, and that these estimates can improve coordination when behaviour is largely determined by feasibility constraints. In \texttt{TidyUP}, CE-CM rapidly converges to the correct capability vector and produces joint plans that closely align with the partner. In contrast, in \texttt{Overcooked}, although CE-CM still learns meaningful approximations of the capability vector, these estimates do not always translate into improved coordination. CE-CM-Div addresses this limitation, enabling robust capability inference from human demonstrations where behaviour is highly variable.

At the same time, our experiments reveal a key insight: correctly estimating a partner’s capabilities is not sufficient to predict their behaviour. Capabilities constrain which plans are feasible, but they do not determine which plan will be executed. In domains such as \texttt{Overcooked}, where multiple joint behaviours can achieve similar outcomes, the absence of strong preferences leads to ambiguity that cannot be resolved through capability inference alone. This explains why improved capability estimates do not necessarily reduce coordination errors in such settings.

CE-CM-Div mitigates this limitation by reasoning over a set of plausible trajectories for each capability hypothesis. By matching observations against multiple rollouts rather than a single predicted trajectory, it becomes significantly more robust to behavioural variability and suboptimality, as demonstrated in the human-subject study. However, CE-CM-Div still does not explicitly model preferences, and therefore cannot systematically resolve ambiguity between equally feasible behaviours. In addition, it incurs higher computational cost due to the need for diverse planning.

A further limitation of both CE-CM and CE-CM-Div is scalability beyond dyadic interactions. Joint planning scales exponentially with the number of agents, making direct extensions to larger teams impractical. Addressing this limitation will likely require structured approximations, such as decomposed planning, role-based reasoning, or factorised representations of multi-agent interactions.

Finally, a promising direction for future work is to jointly model capabilities and preferences. Such an approach would allow agents to distinguish between feasibility constraints and behavioural tendencies, enabling more accurate prediction of partner behaviour. Incorporating preference inference, for example through hierarchical or layered partner models \cite{albrecht2019Reasoning}, could significantly improve coordination in settings where multiple valid strategies exist.

\section{Conclusion}

In this paper, we introduced CE-CM, a framework for task-agnostic adaptation in ad-hoc teamwork based on inferring latent partner capabilities. By modelling capabilities as task-invariant constraints on feasible behaviour, CE-CM provides an explicit and interpretable representation of what a partner can and cannot do, enabling agents to refine their beliefs across tasks and coordinate through capability-conditioned planning.

Our experiments show that CE-CM can reliably recover partner capabilities in well-specified settings and use these estimates to improve coordination by avoiding infeasible or unsafe action assignments. However, we also demonstrate a key limitation: capabilities alone are not sufficient to predict behaviour when multiple valid strategies exist. In such settings, coordination depends not only on feasibility but also on preferences and conventions.

To address this, we introduced CE-CM-Div, which extends capability inference by reasoning over sets of plausible behaviours rather than single trajectories. This significantly improves robustness to behavioural variability, particularly in human–AI teaming scenarios where observed behaviour is diverse and often suboptimal.

Overall, our results highlight capability inference as a principled mechanism for generalising across tasks and partners, while also emphasising the importance of modelling behavioural diversity in realistic settings. Future work includes integrating preference inference alongside capabilities, improving computational efficiency for real-time deployment, and extending the approach to larger multi-agent teams.

\newpage

\appendix

\section{Domain Descriptions} %\label{app:domains}

In this section, we provide detailed descriptions of the two experimental domains that we used to evaluate CE-CM and CE-CM-Div.

\subsection{TidyUP} \label{app:domains-tu}

\texttt{TidyUP} is a PDDL-based human–robot collaboration domain representing a household environment. The domain consists of:

\begin{itemize}
\item \textbf{Rooms:} $\{kitchen, living\_room, bedroom, study, bathroom\}$
\item \textbf{Objects:} $\{plate, bowl, toothbrush, blanket\}$
\item \textbf{Object states:} $\{dirty, clean\}$
\item \textbf{Actions:} $\{move, pick, place, wash, no\_op\}$
\end{itemize}

States are represented as binary predicate vectors encoding object locations, object states, and agent positions. Joint actions are defined as $\mathcal{A} = A \times A \setminus \{(no\_op, no\_op)\}$.

The objective is to reach a goal state in which all objects are clean and placed in designated rooms. Action costs induce preferences over joint plans:
\begin{itemize}
\item The human prefers handling plates and bowls (cost 1 vs. 4 for the robot),
\item Slight preference for toothbrush (1 vs. 2),
\item No preference for blanket (cost 1 for both).
\end{itemize}

Transitions are deterministic and governed by action preconditions. A transition occurs only if all required conditions and capabilities are satisfied; otherwise, the action results in a self-loop. For example, picking up an object requires being in the same room, having free hands, and possessing the corresponding capability.

The domain contains 20 binary capabilities, summarised in Table~\ref{tab:caps}. Some capabilities (e.g., washing outside the kitchen) are inactive due to environmental constraints and do not affect behaviour.

The robot agent is fully capable. The human agent’s capability vector varies across experiments.

\begin{table}[h]
\centering
\scriptsize
\begin{tabular}{ll} 
\toprule
\textbf{Capability} & \textbf{Description} \\
\midrule
$place\_living\_room$ & Ability to put down objects in the living room \\
$place\_kitchen$ & Ability to put down objects in the kitchen \\
$place\_bedroom$ & Ability to put down objects in the bedroom \\
$place\_bathroom$ & Ability to put down objects in the bathroom \\
$place\_study$ & Ability to put down objects in the kitchen \\
$pick\_living\_room$ & Ability to pick up objects in the living room \\
$pick\_kitchen$ & Ability to pick up objects in the kitchen \\
$pick\_bedroom$ & Ability to pick up objects in the bedroom \\
$pick\_bathroom$ & Ability to pick up objects in the bathroom \\
$pick\_study$ & Ability to pick up objects in the kitchen \\
$move\_living\_room$ & Ability to go to the living room \\
$move\_kitchen$ & Ability to go to the kitchen \\
$move\_bedroom$ & Ability to go to the bedroom \\
$move\_bathroom$ & Ability to go to the bathroom \\
$move\_study$ & Ability to go to the kitchen \\
$wash\_living\_room$ & Ability to clean objects in the living room \\
$wash\_kitchen$ & Ability to clean objects in the kitchen \\
$wash\_bedroom$ & Ability to clean objects in the bedroom \\
$wash\_bathroom$ & Ability to clean objects in the bathroom \\
$wash\_study$ & Ability to clean objects in the kitchen \\
\bottomrule
\end{tabular}
\caption{List of capabilities in the \texttt{TidyUP} environment.} \label{tab:caps}
\end{table}

\subsection{Overcooked} \label{app:domains-oc}

We use a macro-action version of \texttt{Overcooked} based on \cite{xiao2025asynchronous}. Macro actions abstract low-level movements into high-level commands. The individual movement and interaction commands are produced by an $A^*$ planner that, given a macro action as a goal, produces a sequence of primitive actions that execute the macro action.

The domain consists of:

\begin{itemize}
    \item \textbf{Agents:} $\{blue\_agent, green\_agent\}$
    \item \textbf{Objects:} $\{lettuce, onion, tomato, plate_1, plate_2, knife_1, knife_2, delivery\}$
    \item \textbf{Vegetable states:} $\{raw, chopped\}$
    \item \textbf{Actions:}
    \begin{equation*}
        \begin{split}
        \{stay, get\_lettuce, get\_tomato, get\_onion, get\_plate\_1, get\_plate\_2, \\ 
        go\_to\_knife\_1, go\_to\_knife\_2, deliver, chop, out\_of\_the\_way\}.
        \end{split}
    \end{equation*}
\end{itemize}

The state representation combines binary and continuous features encoding object positions, agent positions, and task information. Joint actions are defined as $\mathcal{A} = A \times A \setminus \{(stay, stay)\}$. Transitions are deterministic and conditioned on capabilities: actions that require unavailable capabilities result in self-loops. The domain contains 10 binary capabilities (Table~\ref{tab:caps-oc}), defining whether the agent can interact with specific objects or perform specific operations.

The reward structure is sparse:
\begin{itemize}
\item $-0.1$ per step,
\item $+0.5$ for correct chopping,
\item $+1$ for delivering the correct dish,
\item $-1$ for incorrect delivery.
\end{itemize}

We generate 9 tasks using 3 layouts and 3 recipes.

Transitions are deterministic and conditioned on capabilities: actions that require unavailable capabilities result in self-loops.

The domain contains 10 binary capabilities (Table~\ref{tab:caps-oc}), defining whether the agent can interact with specific objects or perform specific operations.

The blue agent (intelligent agent) is fully capable, while the green agent’s capability vector varies across experiments.

\begin{table}[h]
\centering
\scriptsize
\begin{tabular}{ll} 
\toprule
\textbf{Capability} & \textbf{Description} \\
\midrule
$handle\_lettuce$ & Ability to handle raw lettuce \\
$handle\_onion$ & Ability to handle raw onion \\
$handle\_tomato$ & Ability to handle raw tomato \\
$handle\_plate\_1$ & Ability to handle plate 1 \\
$handle\_plate\_2$ & Ability to handle plate 2 \\
$handle\_knife\_1$ & Ability to use knife 1 \\
$handle\_knife\_2$ & Ability to use knife 2 \\
$deliver$ & Ability to deliver a dish \\
$chop$ & Ability to chop vegetables \\
$out\_of\_the\_way$ & Ability to move to the middle of the kitchen \\
\bottomrule
\end{tabular}
\caption{List of capabilities in the \texttt{Overcooked} environment.} \label{tab:caps-oc}
\end{table}

\section{Methods and Baselines Implementation Details} \label{app:methods}

This appendix provides implementation details for CE-CM, CE-CM-Div, and all baselines. The main differences between implementations arise from the choice of planner, similarity metric, and sampling parameters.

\subsection{CE-CM with PDDL} \label{app:methods-tu}

We use the POPF planner \cite{coles2010forward} to generate joint plans in \texttt{TidyUP}. At each iteration, we sample $N=500$ capability vectors from a Bernoulli prior with $P(c_i=1)=0.8$.

Trajectory similarity is measured using Jaccard distance over binary state representations. We set the acceptance threshold to $\varepsilon=0.24$. The capability estimate is obtained by thresholding the Bernoulli parameters with $\psi=0.5$.

The optimistic baseline assumes full capabilities and performs no learning. The pessimistic baseline updates capabilities only when observed, using a cumulative OR over inferred capabilities $\hat{c}_{n-1}$ to update the belief $\hat{c}_{n}=\hat{c}_{n-1} \oplus c_{obs}$.

\subsection{CE-CM with MCTS} \label{app:methods-oc}

In \texttt{Overcooked}, we implement CE-CM using PUCT MCTS \cite{silver2017mastering}. The planner is biased by a pre-trained joint-action policy $\pi^*_{pt}$, yielding the selection rule:

\[
a^*=\argmax_{a \in A}\left[\hat{Q}(s,a)+c_p \pi^*_{pt}(a|s)\cdot \frac{\sqrt{n(s)}}{1+N(s,a)}\right],
\]
with $c_p=1$.

We sample $N=500$ capability vectors from the full space ($2^{10}$). For efficiency, rollout trajectories are precomputed and reused across tasks. We also assume access to a function $f:T \times A \rightarrow \{0,1\}$, which prunes infeasible actions from the planner's consideration. We do so to reduce the branching factor of the joint planner and stop it from considering unproductive actions and help it find feasible joint plans quicker.

Trajectory similarity is measured using cosine similarity over flattened state vectors, with threshold $\varepsilon=0.03$.

Table~\ref{tab:mcts} summarises MCTS hyperparameters.

\begin{table}[h!]
\centering
\scriptsize
\begin{tabular}{lc} 
\toprule
\textbf{Parameter} & \textbf{Value} \\
\midrule
Number of simulations & 200 \\
Number of leaf node rollouts & 12 \\
Maximum rollout depth & 20 \\
Discount factor $\gamma$ & 0.99 \\
\bottomrule
\end{tabular}
\caption{Hyperparameters for the PUCT MCTS planner.}
\label{tab:mcts}
\end{table}

\subsection{Q-learning Baseline}

We implement the optimistic baseline using a tabular Q-learning policy assuming a fully capable partner. The policy is trained for 70M steps using softmax exploration and replay. Other hyperparameters can be found in Table~\ref{tab:params} below.

At inference time, actions are selected using a softmax policy with temperature $\tau=0.1$.

We note here that the behaviour of the Q-policy-based agent approximates to one that uses MCTS to plan its actions assuming a fully-capable partner when the Q-policy has converged. We make this assumption in our evaluation, and thus do not compare to an MCTS agent in our experiments.

\begin{table}[h!]
\centering
\scriptsize
\begin{tabular}{lc} 
\toprule
\textbf{Parameter} & \textbf{Value} \\
\midrule
No. of training steps & 70M \\
Initial $Q$ values & 0.01 \\
Exploration mechanism & Softmax / Random (0.8 / 0.2) \\
Softmax policy temperature & 1.0 \\
Replay buffer size & 100k \\
Replay buffer update frequency & 100k \\
Learning rate $\alpha$ & 0.1 \\
Discount factor $\gamma$ & 0.99 \\
\bottomrule
\end{tabular}
\caption{Hyperparameters used for the baseline policy.}
\label{tab:params}
\end{table}

\subsection{CE-CM-Div} \label{app:ce-cm-div}

CE-CM-Div extends CE-CM by evaluating each capability hypothesis against a set of trajectories $\Delta$ rather than a single rollout. In our implementation, we generate the set of trajectories $\Delta$ using a diversity-aware MCTS procedure \cite{benke2023diverse}.

Candidate trajectories are extracted from the MCTS search tree and filtered to ensure diversity. Diversity is measured using trajectory set difference \cite{srivastava2007domain}:

\[
\textsc{Diversity}(\tau, \Gamma) = \frac{1}{|\Gamma|} \min_{\tau' \in \Gamma} \left[ \frac{|S_{\tau} \setminus S_{\tau'}|}{|S_{\tau}|} \right].
\]

We retain up to $k=50$ trajectories with diversity above threshold $\delta=0.35$. Table~\ref{tab:div} summarises the hyperparameters.

Algorithm~\ref{alg:extract-trajs} describes the extraction procedure.

We emphasise that CE-CM-Div is not tied to this implementation; any method capable of generating diverse trajectories can be used.

\begin{table}[h!]
\centering
\scriptsize
\begin{tabular}{lc} 
\toprule
\textbf{Parameter} & \textbf{Value} \\
\midrule
Maximum no. of trajectories $k$ & 50 \\
Diversity parameter & 0.35 \\
Minimum trajectory length &  \\
Maximum trajectory length & Tree depth \\
\bottomrule
\end{tabular}
\caption{Hyperparameters used for the breadth-first search algorithm.}
\label{tab:div}
\end{table}

We emphasise that CE-CM-Div is not tied to this specific implementation. Any planning method capable of generating multiple diverse trajectories under a given capability hypothesis can be used in place of DiverseMCTS. The choice of planner and diversity metric may affect performance, but does not change the underlying inference procedure.

\begin{algorithm}[htpb]
\scriptsize
\caption{DiverseMCTS}
\label{alg:extract-trajs}
\KwIn{Task $g$, capability estimate $c$, minimum length $L_{\min}$, diversity threshold $\delta$, maximum number of trajectories $k$}
\KwOut{Diverse trajectory set $\Delta = \langle \tau_1, \dots, \tau_k \rangle$}

$root \gets \textsc{MCTS}(g, c)$\;
$\Gamma \gets \emptyset$\;
$\mathit{open\_list} \gets [[root]]$\;

\While{$\mathit{open\_list}$}{
    $p \gets \mathit{open\_list}.pop()$\;
    $n \gets p.last()$\;
    $L \gets |p| - 1$\;
    
    \eIf{$L \geq d_{\max}$ \textbf{or} $n.children = \emptyset$}{
        \If{$L \geq L_{\min}$}{
            $\Gamma \gets \Gamma \cup p$\;
        }
    }{
        \ForEach{child $c \in n.children$}{
            $\mathit{open\_list} \gets \mathit{open\_list} \cup [p.extend(c)]$\;
        }
    }
}

Sort $\Gamma$ by descending length\;
$\Delta \gets \emptyset$\;

\ForEach{$p \in \Gamma$}{
    \If{$|\Delta| \geq k$}{
        \textbf{break}\;
    }
    $d \gets \textsc{Diversity}(p, \Gamma)$\;
    \If{$d \geq \delta$}{
        $\Delta \gets \Delta \cup p$\;
    }
}

\Return $\Delta$
\end{algorithm}

\section{Additional Results}

This section provides supplementary results that support the findings presented in Section~\ref{sec:exp}. These experiments offer additional insight into (i) how capability estimates evolve over time, (ii) how CE-CM adapts to changing capabilities, and (iii) how the diversity parameter $\delta$ affects the performance of CE-CM-Div. While these results are not central to the main claims of the paper, they provide further evidence of the behaviour and limitations of our methods.

\subsection{Evolution of Capability Estimates}

\begin{figure*}[h!]
    \centering
    \begin{subfigure}{\linewidth}
        \centering
        \includegraphics[width=0.85\textwidth]{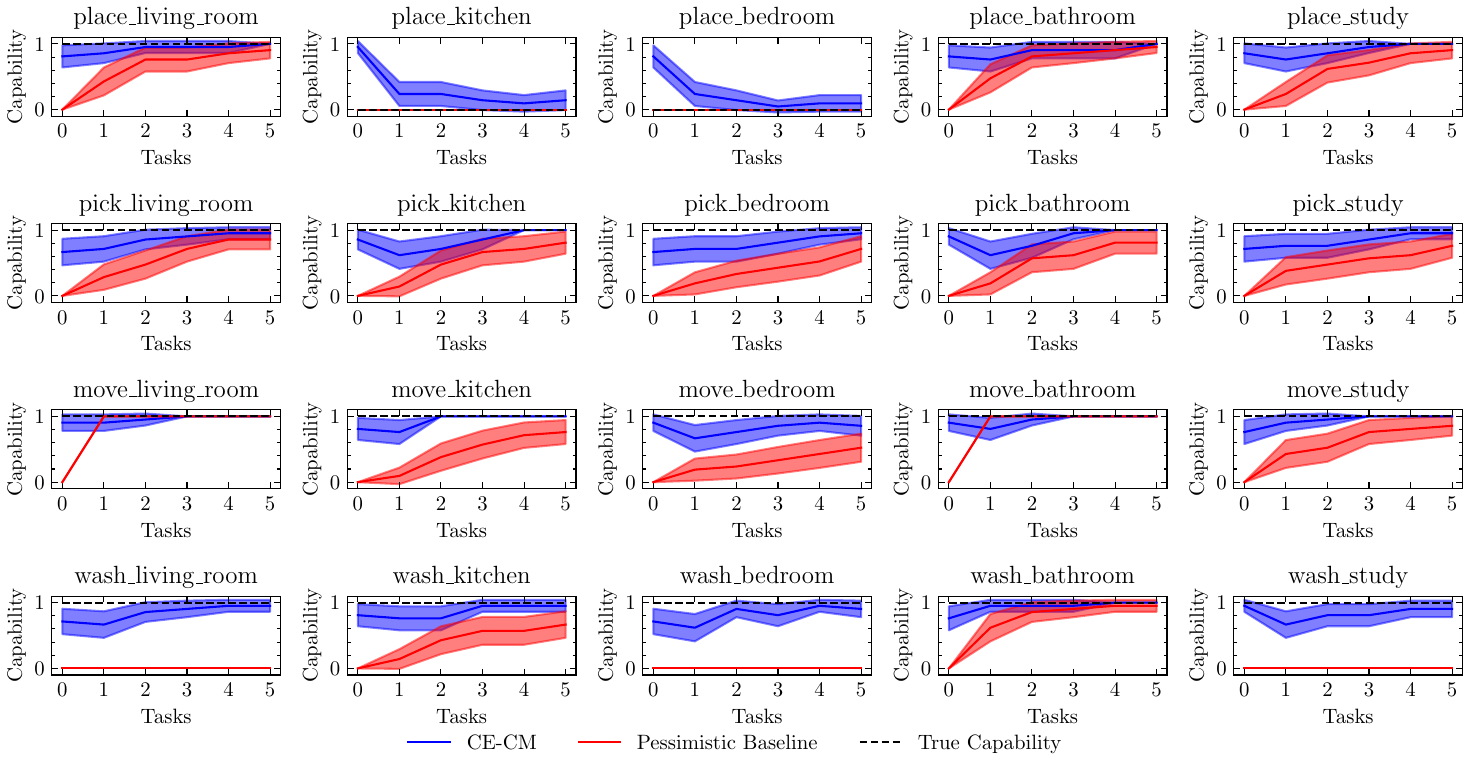}
        \caption{HC Type}
        \label{fig:capabilities_learning_hc}
    \end{subfigure}

    \begin{subfigure}{\linewidth}
        \centering
        \includegraphics[width=0.85\textwidth]{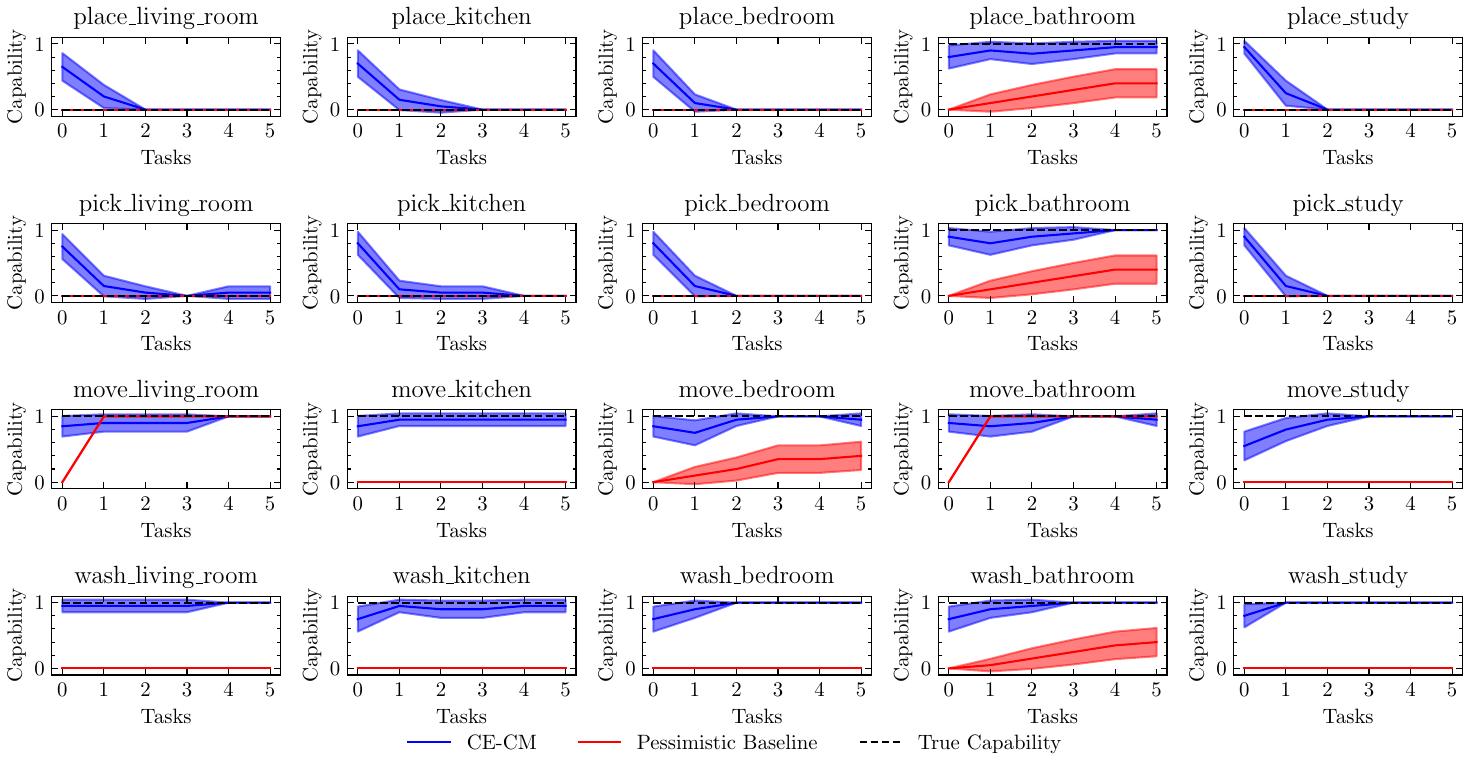}
        \caption{LC Type}
        \label{fig:capabilities_learning_lc}
    \end{subfigure}
    \caption{Evolution of individual capability estimates across tasks. Solid lines show the mean over 20 runs, and shaded regions indicate 95\% confidence intervals.}
\end{figure*}

We visualise the evolution of individual capability estimates across tasks in \texttt{TidyUP}. CE-CM converges to the correct capability values more rapidly than the pessimistic baseline, particularly when multiple capabilities must be inferred jointly. This highlights the advantage of reasoning over the full capability vector rather than updating each capability independently.

\subsection{Adaptation to Changing Capabilities}

We evaluate CE-CM in a setting where the partner’s capabilities change during interaction. The method successfully abandons outdated beliefs and converges to the new capability profile, reducing Hamming distance to approximately $0.1$. This demonstrates that maintaining a distribution over capability hypotheses enables rapid adaptation when the underlying partner model changes.

\begin{figure}[htpb]
    \centering
    \includegraphics[width=0.6\linewidth]{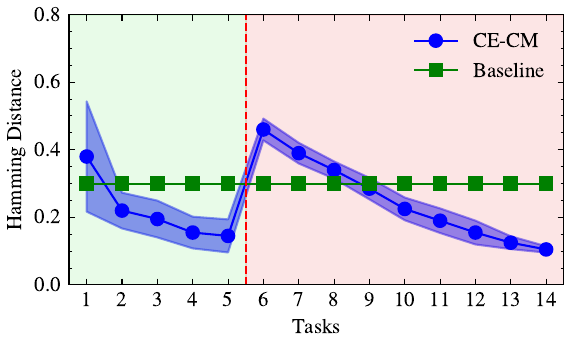}
    \caption{Adaptation to changing capabilities. The partner switches from Type 2 (green) to Type 3 (red) after 5 tasks. Solid line indicates the mean, and shaded area denotes the 95\% Confidence Interval.}
    \label{fig:changing-capabilities}
\end{figure}

\subsection{Effect of the Diversity Parameter in CE-CM-Div} \label{app:diversity}

\begin{figure}[htpb]
    \centering
    \begin{subfigure}[b]{0.4\textwidth}
        \centering
        \includegraphics[width=\textwidth]{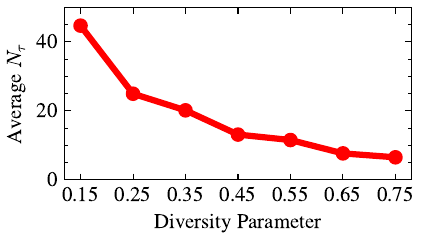}
        \caption{Number of trajectories per capability vector}
        \label{fig:diversity-trajectories}
    \end{subfigure}
    \hfill
    \begin{subfigure}[b]{0.45\textwidth}
        \centering
        \includegraphics[width=\textwidth]{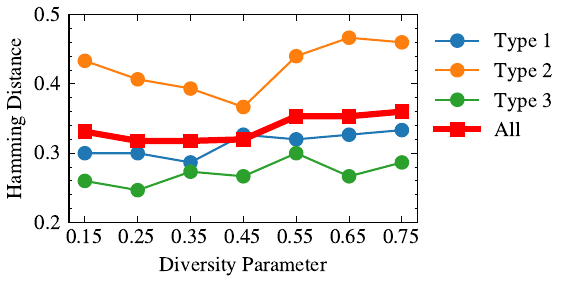}
        \caption{Capability estimation accuracy}
        \label{fig:diversity-hamming}
    \end{subfigure}
    \caption{Impact of the diversity parameter $\delta$ on trajectory diversity and estimation performance.}
\end{figure}

We analyse the effect of the diversity parameter $\delta$ in CE-CM-Div. Increasing $\delta$ reduces the number of trajectories retained per capability vector, as stricter diversity constraints filter out more behaviours (Figure~\ref{fig:diversity-trajectories}). 

Estimation performance exhibits a non-monotonic relationship with $\delta$ (Figure~\ref{fig:diversity-hamming}). Moderate values yield the best results, while too little diversity fails to capture behavioural variation and too much diversity removes informative trajectories. This highlights a trade-off between coverage and selectivity in diverse inference.

\section{Human Study Dataset Collection} \label{app:data-collection}

We collected a dataset of human–agent interactions in the \texttt{Overcooked} domain from 15 participants (8 male, 7 female; mean age $28.27 \pm 3.15$). Participants reported an average video game experience of $3.00 \pm 1.07$ on a 5-point Likert scale (1 = no experience, 5 = daily play). All participants provided informed consent prior to the study.

Each participant completed 15 episodes, organised into three sets of five tasks corresponding to the three capability types used in the main experiments. Tasks were randomly sampled from the 9 layout–recipe combinations described in Section~\ref{sec:exp}. In each episode, participants collaborated with a fixed-policy assistant agent Robbie, controlled by a Q-learning-based joint policy trained with fully capable agents (see~\ref{app:methods-oc}).

Participants selected high-level actions (e.g., \emph{get ingredient}, \emph{chop}, \emph{deliver}) via keyboard input. At each decision point, Robbie’s intended action was displayed, and participants could override it if desired. This interaction protocol mirrors the correction-based execution model used in the simulated experiments.

We recorded sequences of joint states along with the ground-truth capability vector associated with each participant’s assigned type. In total, the dataset contains 225 trajectories (75 per capability type), which are used for offline capability inference in Section~\ref{sec:exp-human}.

\section{Human Study Dataset Analysis} \label{app:dataset}

\begin{figure}[htb]
    \centering
    \includegraphics[width=\linewidth]{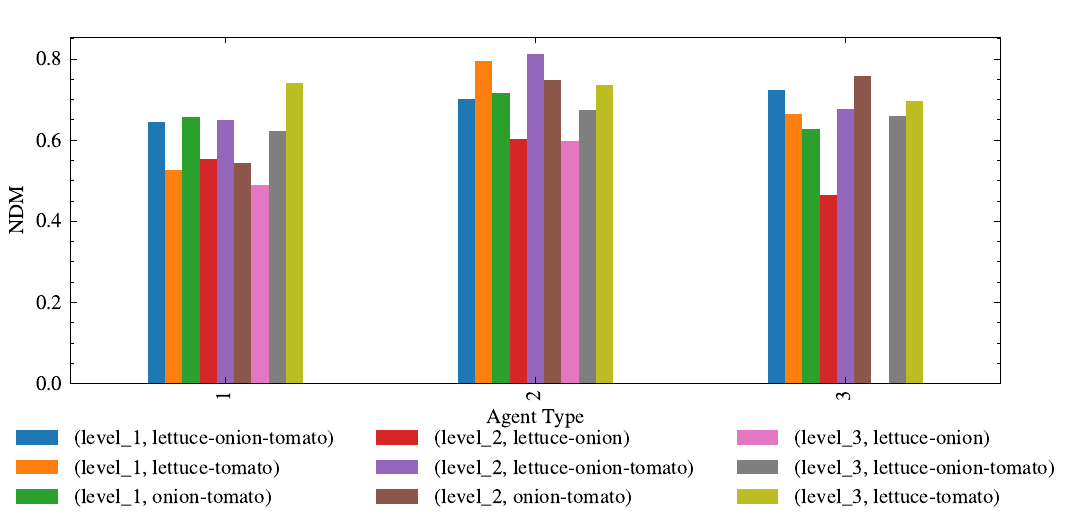}
    \caption{Diversity of the collected trajectories by level and agent type. Each coloured bar represents one task, grouped by agent type.}
    \label{fig:diversity}
\end{figure}

We first examine the diversity of the collected trajectories to evaluate the hypothesis that human behaviour exhibits substantial variability even within the same capability profile. To quantify this, we compute the average pairwise Euclidean distance between trajectory feature vectors, normalised by the maximum distance observed within each task-type group. We refer to this metric as the normalised distance metric (NDM):

\begin{equation*}
    NDM = \frac{1}{N}\sum_{i=1}^N\sum_{j \neq i}\frac{||\tau_i - \tau_j||_2}{D_{max}},
\end{equation*}
where $N$ is the number of trajectories in the group, and $D_{max} = \max_{\tau_1,\tau_2} ||\tau_1 - \tau_2||_2$.

As shown in Figure~\ref{fig:diversity}, trajectory diversity varies significantly across tasks and capability types. Some task–recipe combinations exhibit near-deterministic behaviour (e.g., level\_3 with lettuce-onion), where all participants converge to the same strategy. In contrast, more complex tasks (e.g., level\_2 and level\_3 with lettuce-onion-tomato) exhibit high diversity, indicating multiple viable strategies even under identical capabilities.

\begin{figure}[t]
    \centering
    \includegraphics[width=\linewidth]{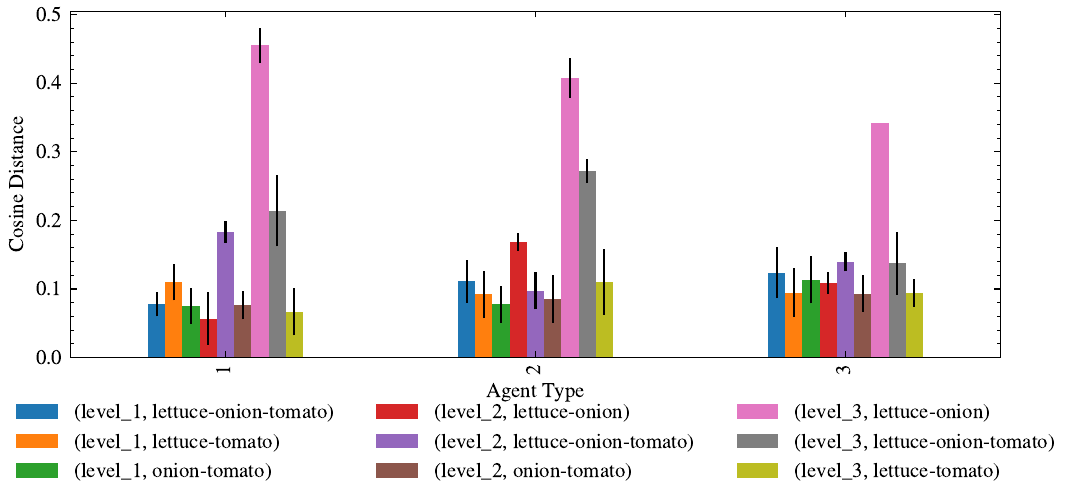}
    \caption{Cosine distance between human trajectories and planner-generated trajectories. Bars show the mean distance, with standard deviation indicated in black.}
    \label{fig:distances}
\end{figure}

Next, we compare human trajectories to those generated by the MCTS planner under the same task and capability conditions. We compute cosine distance between flattened trajectory feature vectors, consistent with the similarity metric used in CE-CM. As shown in Figure~\ref{fig:distances}, planner-generated trajectories differ substantially from human behaviour across all capability types, with considerable variability within each group.

This result highlights a key limitation of single-trajectory inference: even when the capability vector is correct, the planner’s trajectory is unlikely to match the observed human trajectory. Consequently, using a single ``optimal'' rollout provides a weak learning signal for capability inference. This motivates the use of CE-CM-Div, which accounts for behavioural variability by evaluating each capability hypothesis against a set of plausible trajectories rather than a single prediction.

%\nocite{*}
\begin{NoHyper}    
\bibliographystyle{elsarticle-num} 
\bibliography{ce-cm-refs}

@article{albrecht2016Belief,
  title = {Belief and Truth in Hypothesised Behaviours},
  author = {Albrecht, Stefano V. and Crandall, Jacob W. and Ramamoorthy, Subramanian},
  year = {2016},
  journal = {Artificial Intelligence},
  shortjournal = {Artificial Intelligence},
  volume = {235},
  pages = {63--94},
  issn = {00043702},
  doi = {10.1016/j.artint.2016.02.004},
  langid = {english}
}

@article{albrecht2018Autonomous,
  title = {Autonomous Agents Modelling Other Agents: {{A}} Comprehensive Survey and Open Problems},
  author = {Albrecht, Stefano V. and Stone, Peter},
  year = {2018},
  journal = {Artificial Intelligence},
  shortjournal = {Artificial Intelligence},
  volume = {258},
  pages = {66--95},
  issn = {00043702},
  doi = {10.1016/j.artint.2018.01.002},
  langid = {english}
}

@inproceedings{albrecht2019Reasoning,
  title = {Reasoning about {{Hypothetical Agent Behaviours}} and Their {{Parameters}}},
  author = {Albrecht, Stefano V. and Stone, Peter},
  year = {2019},
  eprint = {1906.11064},
  eprinttype = {arXiv},
  eprintclass = {cs},
  doi = {10.48550/arXiv.1906.11064},
  langid = {english},
  booktitle = {Proceedings of the 16th Conference on Autonomous Agents and MultiAgent Systems},
  pages = {547–555},
  numpages = {9},
}

@article{ali2022Heterogeneous,
  title = {Heterogeneous Human–Robot Task Allocation Based on Artificial Trust},
  author = {Ali, Arsha and Azevedo-Sa, Hebert and Tilbury, Dawn M. and Robert, Lionel P.},
  year = {2022},
  journal = {Scientific Reports},
  shortjournal = {Sci Rep},
  volume = {12},
  number = {1},
  pages = {15304},
  issn = {2045-2322},
  doi = {10.1038/s41598-022-19140-5},
  langid = {english}
}

@article{barrett2015Cooperating,
  title = {Cooperating with {{Unknown Teammates}} in {{Complex Domains}}: {{A Robot Soccer Case Study}} of {{Ad Hoc Teamwork}}},
  author = {Barrett, Samuel and Stone, Peter},
  date = {2015-02-18},
  journal = {Proceedings of the AAAI Conference on Artificial Intelligence},
  shortjournal = {AAAI},
  volume = {29},
  number = {1},
  issn = {2374-3468, 2159-5399},
  doi = {10.1609/aaai.v29i1.9428},
  langid = {english},
  year = {2015}
}

@article{barrett2017Making,
  title = {Making Friends on the Fly: {{Cooperating}} with New Teammates},
  author = {Barrett, Samuel and Rosenfeld, Avi and Kraus, Sarit and Stone, Peter},
  date = {2017-01-01},
  journal = {Artificial Intelligence},
  shortjournal = {Artificial Intelligence},
  volume = {242},
  pages = {132--171},
  issn = {0004-3702},
  doi = {10.1016/j.artint.2016.10.005},
  langid = {english},
  year = {2017}
}

@article{beaumont2019approximate,
  title={Approximate bayesian computation},
  author={Beaumont, Mark A},
  journal={Annual review of statistics and its application},
  volume={6},
  number={1},
  pages={379--403},
  year={2019},
  publisher={Annual Reviews}
}

@incollection{benke2023Diverse,
  title = {Diverse, {{Top-k}}, and {{Top-Quality Planning Over Simulators}}},
  booktitle = {Frontiers in {{Artificial Intelligence}} and {{Applications}}},
  author = {Benke, Lyndon and Miller, Tim and Papasimeon, Michael and Lipovetzky, Nir},
  editor = {Gal, Kobi and Nowé, Ann and Nalepa, Grzegorz J. and Fairstein, Roy and Rădulescu, Roxana},
  date = {2023-09-28},
  publisher = {IOS Press},
  doi = {10.3233/FAIA230275},
  isbn = {978-1-64368-436-9 978-1-64368-437-6},
  langid = {english},
  pages = {1-10},
  year = {2023}
}

@inproceedings{bestick2018Learning,
  title = {Learning {{Human Ergonomic Preferences}} for {{Handovers}}},
  booktitle = {2018 {{IEEE International Conference}} on {{Robotics}} and {{Automation}} ({{ICRA}})},
  author = {Bestick, Aaron and Pandya, Ravi and Bajcsy, Ruzena and Dragan, Anca D.},
  year = {2018},
  pages = {3257--3264},
  publisher = {IEEE},
  location = {Brisbane, QLD},
  doi = {10.1109/ICRA.2018.8461216},
  eventtitle = {2018 {{IEEE International Conference}} on {{Robotics}} and {{Automation}} ({{ICRA}})},
  isbn = {978-1-5386-3081-5},
  langid = {english}
}

@article{canal2019Adapting,
  title = {Adapting Robot Task Planning to User Preferences: An Assistive Shoe Dressing Example},
  author = {Canal, Gerard and Alenyà, Guillem and Torras, Carme},
  year = {2019},
  journal = {Autonomous Robots},
  shortjournal = {Auton Robot},
  volume = {43},
  number = {6},
  pages = {1343--1356},
  issn = {0929-5593, 1573-7527},
  doi = {10.1007/s10514-018-9737-2},
  langid = {english}
}

@article{carroll2019Utility,
  title={On the utility of learning about humans for human-ai coordination},
  author={Carroll, Micah and Shah, Rohin and Ho, Mark K and Griffiths, Tom and Seshia, Sanjit and Abbeel, Pieter and Dragan, Anca},
  journal={Advances in neural information processing systems},
  volume={32},
  year={2019}
}

@inproceedings{czechowski2020Decentralized,
  title = {Decentralized {{MCTS}} via {{Learned Teammate Models}}},
  booktitle = {Proceedings of the {{Twenty-Ninth International Joint Conference}} on {{Artificial Intelligence}}},
  author = {Czechowski, Aleksander and Oliehoek, Frans A.},
  year = {2020},
  pages = {81--88},
  publisher = {International Joint Conferences on Artificial Intelligence Organization},
  location = {Yokohama, Japan},
  doi = {10.24963/ijcai.2020/12},
  eventtitle = {Twenty-{{Ninth International Joint Conference}} on {{Artificial Intelligence}} and {{Seventeenth Pacific Rim International Conference}} on {{Artificial Intelligence}} \{{{IJCAI-PRICAI-20}}\}},
  isbn = {978-0-9992411-6-5},
  langid = {english}
}

@inproceedings{emam2020Adaptive,
  title = {Adaptive {{Task Allocation}} for {{Heterogeneous Multi-Robot Teams}} with {{Evolving}} and {{Unknown Robot Capabilities}}},
  booktitle = {2020 {{IEEE International Conference}} on {{Robotics}} and {{Automation}} ({{ICRA}})},
  author = {Emam, Yousef and Mayya, Siddharth and Notomista, Gennaro and Bohannon, Addison and Egerstedt, Magnus},
  year = {2020},
  pages = {7719--7725},
  publisher = {IEEE},
  location = {Paris, France},
  doi = {10.1109/ICRA40945.2020.9197283},
  eventtitle = {2020 {{IEEE International Conference}} on {{Robotics}} and {{Automation}} ({{ICRA}})},
  isbn = {978-1-7281-7395-5},
  langid = {english}
}

@article{fu2023Robust,
  title = {Robust {{Task Scheduling}} for {{Heterogeneous Robot Teams Under Capability Uncertainty}}},
  author = {Fu, Bo and Smith, William and Rizzo, Denise M. and Castanier, Matthew and Ghaffari, Maani and Barton, Kira},
  date = {2023-04},
  journal = {IEEE Transactions on Robotics},
  shortjournal = {IEEE Trans. Robot.},
  volume = {39},
  number = {2},
  pages = {1087--1105},
  issn = {1552-3098, 1941-0468},
  doi = {10.1109/TRO.2022.3216068},
  langid = {english},
  year = {2023}
}

@article{ghosh2020Deployment,
  title = {Towards {{Deployment}} of {{Robust Cooperative AI Agents}}: {{An Algorithmic Framework}} for {{Learning Adaptive Policies}}},
  author = {Ghosh, Ahana and Tschiatschek, Sebastian and Mahdavi, Hamed and Singla, Adish},
  year = {2020},
  journal = {New Zealand},
  langid = {english}
}

@inproceedings{gorur2019Anticipatory,
  title={Anticipatory bayesian policy selection for online adaptation of collaborative robots to unknown human types},
  author={G{\"o}r{\"u}r, O Can and Rosman, Benjamin and Albayrak, Sahin},
  booktitle={Proceedings of the 18th International Conference on Autonomous Agents and MultiAgent Systems},
  pages={77--85},
  year={2019}
}

@unpublished{hallak2015Contextual,
  title = {Contextual {{Markov Decision Processes}}},
  author = {Hallak, Assaf and Di Castro, Dotan and Mannor, Shie},
  date = {2015-02-08},
  eprint = {1502.02259},
  eprinttype = {arXiv},
  eprintclass = {cs, stat},
  url = {http://arxiv.org/abs/1502.02259},
  langid = {english},
  pubstate = {prepublished},
  year = {2015},
  note = {arXiv:1502.02259 [stat.ML]}
}

@article{hiatt2017human,
author = {Laura M Hiatt and Cody Narber and Esube Bekele and Sangeet S Khemlani and J Gregory Trafton},
title ={Human modeling for human–robot collaboration},
journal = {The International Journal of Robotics Research},
volume = {36},
number = {5-7},
pages = {580-596},
year = {2017},
doi = {10.1177/0278364917690592}
}

@inproceedings{huang2016Anticipatory,
  title = {Anticipatory Robot Control for Efficient Human-Robot Collaboration},
  booktitle = {2016 11th {{ACM}}/{{IEEE International Conference}} on {{Human-Robot Interaction}} ({{HRI}})},
  author = {Huang, Chien-Ming and Mutlu, Bilge},
  year = {2016},
  pages = {83--90},
  publisher = {IEEE},
  location = {Christchurch, New Zealand},
  doi = {10.1109/HRI.2016.7451737},
  eventtitle = {2016 11th {{ACM}}/{{IEEE International Conference}} on {{Human-Robot Interaction}} ({{HRI}})},
  isbn = {978-1-4673-8370-7},
  langid = {english}
}

@INPROCEEDINGS{izquierdo2022improved,
  author={Izquierdo-Badiola, Silvia and Canal, Gerard and Rizzo, Carlos and Alenyà, Guillem},
  booktitle={2022 International Conference on Robotics and Automation (ICRA)}, 
  title={Improved Task Planning through Failure Anticipation in Human-Robot Collaboration}, 
  year={2022},
  volume={},
  number={},
  pages={7875-7880}
}

@inproceedings{laidlaw2025assistancezero,
title={AssistanceZero: Scalably Solving Assistance Games},
author={Cassidy Laidlaw and Eli Bronstein and Timothy Guo and Dylan Feng and Lukas Berglund and Justin Svegliato and Stuart Russell and Anca Dragan},
booktitle={Forty-second International Conference on Machine Learning},
year={2025},
pages = {1-10},
url={https://openreview.net/forum?id=b9hVMJi0t2}
}

@INPROCEEDINGS{leon2022intuitive,
  author={Felip, Javier and Gonzalez-Aguirre, David and Nachman, Lama},
  booktitle={2022 IEEE/RSJ International Conference on Intelligent Robots and Systems (IROS)}, 
  title={Intuitive \& Efficient Human-robot Collaboration via Real-time Approximate Bayesian Inference}, 
  year={2022},
  volume={},
  number={},
  pages={3093-3099},
  doi={10.1109/IROS47612.2022.9982251}
}

@article{li2021Individualized,
  title = {Individualized {{Mutual Adaptation}} in {{Human-Agent Teams}}},
  author = {Li, Huao and Ni, Tianwei and Agrawal, Siddharth and Jia, Fan and Raja, Suhas and Gui, Yikang and Hughes, Dana and Lewis, Michael and Sycara, Katia},
  date = {2021-12},
  journal = {IEEE Transactions on Human-Machine Systems},
  volume = {51},
  number = {6},
  pages = {706--714},
  issn = {2168-2305},
  doi = {10.1109/THMS.2021.3107675},
  eventtitle = {{{IEEE Transactions}} on {{Human-Machine Systems}}},
  year = {2021}
}

@article{liu2022Coordinating,
  title = {Coordinating {{Human-Robot Teams}} with {{Dynamic}} and {{Stochastic Task Proficiencies}}},
  author = {Liu, Ruisen and Natarajan, Manisha and Gombolay, Matthew C.},
  date = {2022-03-31},
  journal = {ACM Transactions on Human-Robot Interaction},
  shortjournal = {J. Hum.-Robot Interact.},
  volume = {11},
  number = {1},
  pages = {1--42},
  issn = {2573-9522, 2573-9522},
  doi = {10.1145/3477391},
  langid = {english},
  year = {2022}
}

@inproceedings{lou2023pecan,
    author = {Lou, Xingzhou and Guo, Jiaxian and Zhang, Junge and Wang, Jun and Huang, Kaiqi and Du, Yali},
    title = {PECAN: Leveraging Policy Ensemble for Context-Aware Zero-Shot Human-AI Coordination},
    year = {2023},
    isbn = {9781450394321},
    publisher = {International Foundation for Autonomous Agents and Multiagent Systems},
    address = {Richland, SC},
    booktitle = {Proceedings of the 2023 International Conference on Autonomous Agents and Multiagent Systems},
    pages = {679–688},
    numpages = {10},
    location = {London, United Kingdom},
    series = {AAMAS '23}
}

@InProceedings{lupu2021trajectory,
  title = 	 {Trajectory Diversity for Zero-Shot Coordination},
  author =       {Lupu, Andrei and Cui, Brandon and Hu, Hengyuan and Foerster, Jakob},
  booktitle = 	 {Proceedings of the 38th International Conference on Machine Learning},
  pages = 	 {7204--7213},
  year = 	 {2021},
  editor = 	 {Meila, Marina and Zhang, Tong},
  volume = 	 {139},
  series = 	 {Proceedings of Machine Learning Research},
  month = 	 {18--24 Jul},
  publisher =    {PMLR}
}

@unpublished{mirsky2022Survey,
  title = {A {{Survey}} of {{Ad Hoc Teamwork Research}}},
  author = {Mirsky, Reuth and Carlucho, Ignacio and Rahman, Arrasy and Fosong, Elliot and Macke, William and Sridharan, Mohan and Stone, Peter and Albrecht, Stefano V.},
  date = {2022-08-16},
  eprint = {2202.10450},
  eprinttype = {arXiv},
  eprintclass = {cs},
  url = {http://arxiv.org/abs/2202.10450},
  langid = {english},
  pubstate = {prepublished},
  year = {2022},
  note = {arXiv:2202.10450 [cs.MA]}
}

@unpublished{narcomey2024Learning,
  title = {Learning {{Human Preferences Over Robot Behavior}} as {{Soft Planning Constraints}}},
  author = {Narcomey, Austin and Tsoi, Nathan and Desai, Ruta and Vázquez, Marynel},
  year = {2024},
  eprint = {2403.19795},
  eprinttype = {arXiv},
  eprintclass = {cs},
  url = {http://arxiv.org/abs/2403.19795},
  langid = {english},
  pubstate = {prepublished},
  note = {arXiv:2403.19795 [cs.RO]}
}

@article{natarajan2023HumanRobot,
  title = {Human-{{Robot Teaming}}: {{Grand Challenges}}},
  author = {Natarajan, Manisha and Seraj, Esmaeil and Altundas, Batuhan and Paleja, Rohan and Ye, Sean and Chen, Letian and Jensen, Reed and Chang, Kimberlee Chestnut and Gombolay, Matthew},
  date = {2023-08-08},
  journal = {Current Robotics Reports},
  shortjournal = {Curr Robot Rep},
  volume = {4},
  number = {3},
  pages = {81--100},
  issn = {2662-4087},
  doi = {10.1007/s43154-023-00103-1},
  langid = {english},
  year = {2023}
}

@article{nguyen2012Generating,
  title = {Generating Diverse Plans to Handle Unknown and Partially Known User Preferences},
  author = {Nguyen, Tuan Anh and Do, Minh and Gerevini, Alfonso Emilio and Serina, Ivan and Srivastava, Biplav and Kambhampati, Subbarao},
  year = {2012},
  journal = {Artificial Intelligence},
  shortjournal = {Artificial Intelligence},
  volume = {190},
  pages = {1--31},
  issn = {00043702},
  doi = {10.1016/j.artint.2012.05.005},
  langid = {english}
}

@inproceedings{nikolaidis2013Humanrobot,
  title = {Human-Robot Cross-Training: {{Computational}} Formulation, Modeling and Evaluation of a Human Team Training Strategy},
  booktitle = {2013 8th {{ACM}}/{{IEEE International Conference}} on {{Human-Robot Interaction}} ({{HRI}})},
  author = {Nikolaidis, Stefanos and Shah, Julie},
  year = {2013},
  pages = {33--40},
  issn = {2167-2148},
  doi = {10.1109/HRI.2013.6483499},
  eventtitle = {2013 8th {{ACM}}/{{IEEE International Conference}} on {{Human-Robot Interaction}} ({{HRI}})}
}

@inproceedings{nikolaidis2015Efficient,
  title = {Efficient {{Model Learning}} from {{Joint-Action Demonstrations}} for {{Human-Robot Collaborative Tasks}}},
  booktitle = {Proceedings of the {{Tenth Annual ACM}}/{{IEEE International Conference}} on {{Human-Robot Interaction}}},
  author = {Nikolaidis, Stefanos and Ramakrishnan, Ramya and Gu, Keren and Shah, Julie},
  year = {2015},
  series = {{{HRI}} '15},
  pages = {189--196},
  publisher = {Association for Computing Machinery},
  location = {New York, NY, USA},
  doi = {10.1145/2696454.2696455},
  isbn = {978-1-4503-2883-8}
}

@article{nikolaidis2015Improved,
  title = {Improved Human–Robot Team Performance through Cross-Training, an Approach Inspired by Human Team Training Practices},
  author = {Nikolaidis, Stefanos and Lasota, Przemyslaw and Ramakrishnan, Ramya and Shah, Julie},
  year = {2015},
  journal = {The International Journal of Robotics Research},
  shortjournal = {The International Journal of Robotics Research},
  volume = {34},
  number = {14},
  pages = {1711--1730},
  issn = {0278-3649, 1741-3176},
  doi = {10.1177/0278364915609673},
  langid = {english}
}

@inproceedings{orlov2022factorial,
author = {Orlov-Savko, Liubove and Jain, Abhinav and Gremillion, Gregory M. and Neubauer, Catherine E. and Canady, Jonroy D. and Unhelkar, Vaibhav},
title = {Factorial Agent Markov Model: Modeling Other Agents' Behavior in presence of Dynamic Latent Decision Factors},
year = {2022},
isbn = {9781450392136},
publisher = {International Foundation for Autonomous Agents and Multiagent Systems},
address = {Richland, SC},
booktitle = {Proceedings of the 21st International Conference on Autonomous Agents and Multiagent Systems},
pages = {982–990},
numpages = {9},
location = {Virtual Event, New Zealand},
series = {AAMAS '22}
}

@inproceedings{raileanu2018modeling,
  title={Modeling Others using Oneself in Multi-Agent Reinforcement Learning},
  author={Roberta Raileanu and Emily L. Denton and Arthur Szlam and Rob Fergus},
  booktitle={International Conference on Machine Learning},
  year={2018},
  pages={4257--4266},
  url={https://api.semanticscholar.org/CorpusID:3622509}
}

@incollection{ribeiro2021Helping,
  title = {Helping {{People}} on the {{Fly}}: {{Ad Hoc Teamwork}} for {{Human-Robot Teams}}},
  booktitle = {Progress in {{Artificial Intelligence}}},
  author = {Ribeiro, João G. and Faria, Miguel and Sardinha, Alberto and Melo, Francisco S.},
  editor = {Marreiros, Goreti and Melo, Francisco S. and Lau, Nuno and Lopes Cardoso, Henrique and Reis, Luís Paulo},
  year = {2021},
  volume = {12981},
  pages = {635--647},
  publisher = {Springer International Publishing},
  location = {Cham},
  doi = {10.1007/978-3-030-86230-5_50},
  isbn = {978-3-030-86229-9 978-3-030-86230-5},
  langid = {english}
}

@article{ribeiro2024HOTSPOT,
  title = {{{HOTSPOT}}: {{An}} Ad Hoc Teamwork Platform for Mixed Human-Robot Teams},
  author = {Ribeiro, João G. and Henriques, Luis Müller and Colcher, Sérgio and Duarte, Julio Cesar and Melo, Francisco S. and Milidiú, Ruy Luiz and Sardinha, Alberto},
  editor = {Cruz-Villar, Carlos Alberto},
  year = {2024},
  journal = {PLOS ONE},
  shortjournal = {PLoS ONE},
  volume = {19},
  number = {6},
  pages = {e0305705},
  issn = {1932-6203},
  doi = {10.1371/journal.pone.0305705},
  langid = {english}
}

@inproceedings{sarkar2023diverse,
author = {Sarkar, Bidipta and Shih, Andy and Sadigh, Dorsa},
title = {Diverse conventions for human-AI collaboration},
year = {2023},
publisher = {Curran Associates Inc.},
address = {Red Hook, NY, USA},
booktitle = {Proceedings of the 37th International Conference on Neural Information Processing Systems},
articleno = {1003},
numpages = {25},
location = {New Orleans, LA, USA},
series = {NIPS '23},
pages={23115--23139}
}

@article{shafipouryourdshahi2022Online,
  title = {On-Line Estimators for Ad-Hoc Task Execution: Learning Types and Parameters of Teammates for Effective Teamwork},
  author = {Shafipour Yourdshahi, Elnaz and Do Carmo Alves, Matheus Aparecido and Varma, Amokh and Soriano Marcolino, Leandro and Ueyama, Jó and Angelov, Plamen},
  year = {2022},
  journal = {Autonomous Agents and Multi-Agent Systems},
  shortjournal = {Auton Agent Multi-Agent Syst},
  volume = {36},
  number = {2},
  pages = {45},
  issn = {1387-2532, 1573-7454},
  doi = {10.1007/s10458-022-09571-9},
  langid = {english}
}

@article{shih2021Critical,
  title = {On the {{Critical Role}} of {{Conventions}} in {{Adaptive Human-AI Collaboration}}},
  author = {Shih, Andy and Sawhney, Arjun and Kondic, Jovana and Ermon, Stefano and Sadigh, Dorsa},
  date = {2021-04-06},
  journal = {ICLR 2021},
  eprint = {2104.02871},
  eprinttype = {arXiv},
  url = {http://arxiv.org/abs/2104.02871},
  langid = {english},
  year = {2021}
}

@inproceedings{srivastava2007domain,
author = {Srivastava, Biplav and Nguyen, Tuan A. and Gerevini, Alfonso and Kambhampati, Subbarao and Do, Minh Binh and Serina, Ivan},
title = {Domain independent approaches for finding diverse plans},
year = {2007},
publisher = {Morgan Kaufmann Publishers Inc.},
address = {San Francisco, CA, USA},
booktitle = {Proceedings of the 20th International Joint Conference on Artifical Intelligence},
pages = {2016–2022},
numpages = {7},
location = {Hyderabad, India},
series = {IJCAI'07}
}

@article{stone2010Ad,
  title = {Ad {{Hoc Autonomous Agent Teams}}: {{Collaboration}} without {{Pre-Coordination}}},
  author = {Stone, Peter and Kaminka, Gal and Kraus, Sarit and Rosenschein, Jeffrey},
  date = {2010-07-05},
  journal = {Proceedings of the AAAI Conference on Artificial Intelligence},
  shortjournal = {AAAI},
  volume = {24},
  number = {1},
  pages = {1504--1509},
  issn = {2374-3468, 2159-5399},
  doi = {10.1609/aaai.v24i1.7529},
  langid = {english},
  year = {2010}
}

@inproceedings{strouse2021collaborating,
author = {Strouse, DJ and McKee, Kevin R. and Botvinick, Matt and Hughes, Edward and Everett, Richard},
title = {Collaborating with humans without human data},
year = {2021},
isbn = {9781713845393},
publisher = {Curran Associates Inc.},
address = {Red Hook, NY, USA},
booktitle = {Proceedings of the 35th International Conference on Neural Information Processing Systems},
articleno = {1111},
numpages = {14},
series = {NIPS '21},
pages={14502--14515}
}

@inproceedings{szot2023adaptive,
author = {Szot, Andrew and Jain, Unnat and Batra, Dhruv and Kira, Zsolt and Desai, Ruta and Rai, Akshara},
title = {Adaptive coordination in social embodied rearrangement},
year = {2023},
publisher = {JMLR.org},
booktitle = {Proceedings of the 40th International Conference on Machine Learning},
articleno = {1388},
numpages = {16},
location = {Honolulu, Hawaii, USA},
series = {ICML'23},
pages={33365--33380}
}

@inproceedings{trivedi2018Inverse,
  title = {Inverse {{Learning}} of {{Robot Behavior}} for {{Collaborative Planning}}},
  booktitle = {2018 {{IEEE}}/{{RSJ International Conference}} on {{Intelligent Robots}} and {{Systems}} ({{IROS}})},
  author = {Trivedi, Maulesh and Doshi, Prashant},
  year = {2018},
  pages = {1--9},
  issn = {2153-0866},
  doi = {10.1109/IROS.2018.8593745},
  eventtitle = {2018 {{IEEE}}/{{RSJ International Conference}} on {{Intelligent Robots}} and {{Systems}} ({{IROS}})}
}

@inproceedings{unhelkar2019semi,
  title={Semi-Supervised Learning of Decision-Making Models for Human-Robot Collaboration},
  author={Vaibhav Unhelkar and Shen Li and Julie A. Shah},
  booktitle={Conference on Robot Learning},
  year={2019},
  pages={192--203},
  url={https://api.semanticscholar.org/CorpusID:208175027}
}

@inproceedings{wang2022co,
  title={Co-gail: Learning diverse strategies for human-robot collaboration},
  author={Wang, Chen and P{\'e}rez-D’Arpino, Claudia and Xu, Danfei and Fei-Fei, Li and Liu, Karen and Savarese, Silvio},
  booktitle={Conference on Robot Learning},
  pages={1279--1290},
  year={2022},
  organization={PMLR}
}

@article{wu2021Too,
  title = {Too {{Many Cooks}}: {{Bayesian Inference}} for {{Coordinating Multi}}‐{{Agent Collaboration}}},
  author = {Wu, Sarah A. and Wang, Rose E. and Evans, James A. and Tenenbaum, Joshua B. and Parkes, David C. and Kleiman‐Weiner, Max},
  date = {2021-04},
  journal = {Topics in Cognitive Science},
  shortjournal = {Topics in Cognitive Science},
  volume = {13},
  number = {2},
  pages = {414--432},
  issn = {1756-8757, 1756-8765},
  doi = {10.1111/tops.12525},
  langid = {english},
    year = {2021}
}

@inproceedings{wu2011online,
author = {Wu, Feng and Zilberstein, Shlomo and Chen, Xiaoping},
title = {Online planning for ad hoc autonomous agent teams},
year = {2011},
isbn = {9781577355137},
publisher = {AAAI Press},
booktitle = {Proceedings of the Twenty-Second International Joint Conference on Artificial Intelligence - Volume One},
pages = {439–445},
numpages = {7},
location = {Barcelona, Catalonia, Spain},
series = {IJCAI'11}
}

@inproceedings{xing2021Learning,
  title = {Learning with {{Generated Teammates}} to {{Achieve Type-Free Ad-Hoc Teamwork}}},
  booktitle = {Proceedings of the {{Thirtieth International Joint Conference}} on {{Artificial Intelligence}}},
  author = {Xing, Dong and Liu, Qianhui and Zheng, Qian and Pan, Gang},
  date = {2021-08},
  pages = {472--478},
  publisher = {International Joint Conferences on Artificial Intelligence Organization},
  location = {Montreal, Canada},
  doi = {10.24963/ijcai.2021/66},
  eventtitle = {Thirtieth {{International Joint Conference}} on {{Artificial Intelligence}} \{{{IJCAI-21}}\}},
  isbn = {978-0-9992411-9-6},
  langid = {english},
  year = {2021}
}

@inproceedings{yu2023learning,
  author       = {Chao Yu and
                  Jiaxuan Gao and
                  Weilin Liu and
                  Botian Xu and
                  Hao Tang and
                  Jiaqi Yang and
                  Yu Wang and
                  Yi Wu},
  title        = {Learning Zero-Shot Cooperation with Humans, Assuming Humans Are Biased},
  booktitle    = {The Eleventh International Conference on Learning Representations,
                  {ICLR} 2023, Kigali, Rwanda, May 1-5, 2023},
  publisher    = {OpenReview.net},
  year         = {2023},
  url          = {https://openreview.net/forum?id=TrwE8l9aJzs},
  pages = {1-10}
}

@inproceedings{zhang2020RealTime,
  title = {Real-{{Time Adaptive Assembly Scheduling}} in {{Human-Multi-Robot Collaboration According}} to {{Human Capability}}},
  booktitle = {2020 {{IEEE International Conference}} on {{Robotics}} and {{Automation}} ({{ICRA}})},
  author = {Zhang, Shaobo and Chen, Yi and Zhang, Jun and Jia, Yunyi},
  date = {2020-05},
  pages = {3860--3866},
  publisher = {IEEE},
  location = {Paris, France},
  doi = {10.1109/ICRA40945.2020.9196618},
  eventtitle = {2020 {{IEEE International Conference}} on {{Robotics}} and {{Automation}} ({{ICRA}})},
  isbn = {978-1-7281-7395-5},
  langid = {english},
  year = {2020}
}

@inproceedings{zhao2022Coordination,
  title = {Coordination {{With Humans Via Strategy Matching}}},
  booktitle = {2022 {{IEEE}}/{{RSJ International Conference}} on {{Intelligent Robots}} and {{Systems}} ({{IROS}})},
  author = {Zhao, Michelle and Simmons, Reid and Admoni, Henny},
  date = {2022-10},
  pages = {9116--9123},
  issn = {2153-0866},
  doi = {10.1109/IROS47612.2022.9982277},
  eventtitle = {2022 {{IEEE}}/{{RSJ International Conference}} on {{Intelligent Robots}} and {{Systems}} ({{IROS}})},
  year = {2022}
}

@InProceedings{zhao2023learning,
  title = 	 {Learning Human Contribution Preferences in Collaborative Human-Robot Tasks},
  author =       {Zhao, Michelle D and Simmons, Reid and Admoni, Henny},
  booktitle = 	 {Proceedings of The 7th Conference on Robot Learning},
  pages = 	 {3597--3618},
  year = 	 {2023},
  editor = 	 {Tan, Jie and Toussaint, Marc and Darvish, Kourosh},
  volume = 	 {229},
  series = 	 {Proceedings of Machine Learning Research},
  month = 	 {06--09 Nov},
  publisher =    {PMLR},
}

@article{Zhao2022the,
  title={The Role of Adaptation in Collective Human–AI Teaming},
  author={Michelle D. Zhao and Reid G. Simmons and Henny Admoni},
  journal={Topics in Cognitive Science},
  year={2022},
  volume={17},
  pages={291 - 323}
}

@article{pynadath2002communicative,
  title={The communicative multiagent team decision problem: Analyzing teamwork theories and models},
  author={Pynadath, David V and Tambe, Milind},
  journal={Journal of artificial intelligence research},
  volume={16},
  pages={389--423},
  year={2002}
}

@article{heinrich2016deep,
  title={Deep reinforcement learning from self-play in imperfect-information games},
  author={Heinrich, Johannes and Silver, David},
  journal={arXiv preprint arXiv:1603.01121},
  year={2016}
}

@article{fox2003pddl2,
  title={PDDL2. 1: An extension to PDDL for expressing temporal planning domains},
  author={Fox, Maria and Long, Derek},
  journal={Journal of artificial intelligence research},
  volume={20},
  pages={61--124},
  year={2003}
}

@inproceedings{coles2010forward,
  title={Forward-chaining partial-order planning},
  author={Coles, Amanda and Coles, Andrew and Fox, Maria and Long, Derek},
  booktitle={Proceedings of the International Conference on Automated Planning and Scheduling},
  volume={20},
  pages={42--49},
  year={2010}
}

@article{xiao2025asynchronous,
author = {Yuchen Xiao and Weihao Tan and Joshua Hoffman and Tian Xia and Christopher Amato},
title ={Asynchronous multi-agent deep reinforcement learning under partial observability},
journal = {The International Journal of Robotics Research},
volume = {44},
number = {8},
pages = {1257-1286},
year = {2025},
doi = {10.1177/02783649241306124}
}

@InProceedings{hu2020other,
  title = 	 {“{O}ther-Play” for Zero-Shot Coordination},
  author =       {Hu, Hengyuan and Lerer, Adam and Peysakhovich, Alex and Foerster, Jakob},
  booktitle = 	 {Proceedings of the 37th International Conference on Machine Learning},
  pages = 	 {4399--4410},
  year = 	 {2020},
  editor = 	 {III, Hal Daumé and Singh, Aarti},
  volume = 	 {119},
  series = 	 {Proceedings of Machine Learning Research},
  month = 	 {13--18 Jul},
  publisher =    {PMLR}
}

@article{silver2017mastering,
  title={Mastering the game of Go without human knowledge},
  author={David Silver and Julian Schrittwieser and Karen Simonyan and Ioannis Antonoglou and Aja Huang and Arthur Guez and Thomas Hubert and Lucas baker and Matthew Lai and Adrian Bolton and Yutian Chen and Timothy P. Lillicrap and Fan Hui and L. Sifre and George van den Driessche and Thore Graepel and Demis Hassabis},
  journal={Nature},
  year={2017},
  volume={550},
  pages={354-359},
  url={https://api.semanticscholar.org/CorpusID:205261034}
}

@inproceedings{tisnikar2024probabilistic,
  title={Probabilistic inference of human capabilities from passive observations},
  author={Tisnikar, Peter and Canal, Gerard and Leonetti, Matteo},
  booktitle={2024 IEEE/RSJ International Conference on Intelligent Robots and Systems (IROS)},
  pages={8779--8785},
  year={2024},
  organization={IEEE}
}
\end{NoHyper}



\end{document}